\documentclass[11pt,a4paper]{article}
\usepackage[hyperref]{naaclhlt2019}
\usepackage{times}
\usepackage{latexsym}
\usepackage{graphicx}
\usepackage{color}
\usepackage{multicol}
\usepackage{url}
\usepackage{multirow}
\usepackage{amssymb}
\usepackage{tabularx}
\usepackage{booktabs}

\title{Ranking and Selecting Multi-Hop Knowledge Paths to Better Predict Human Needs}
\aclfinalcopy
\author{Debjit Paul \\
  Research Training Group AIPHES \\
  Institute for Computational Linguistics\\
  Heidelberg University \\
  {\tt paul@cl.uni-heidelberg.de} \\\And
  Anette Frank \\
  Research Training Group AIPHES \\
  Institute for Computational Linguistics \\
  Heidelberg University\\
  {\tt frank@cl.uni-heidelberg.de} \\}

\date{}
\begin{document}
\maketitle
\begin{abstract}

  To 
  make machines 
  better
  understand sentiments, research needs to move from polarity identification to understanding the reasons that underlie the expression of sentiment. 
  Categorizing the goals or needs of humans is one way to explain the expression of sentiment
  in text. Humans 
  are good at 
 understanding 
  situations described in natural language and can easily connect them to the character's psychological needs using commonsense knowledge. We present a novel method to extract, rank, filter and select multi-hop relation paths from a commonsense knowledge resource 
  to interpret the expression of sentiment in terms of their
  underlying human needs.
  We efficiently integrate the acquired 
  knowledge paths in a neural model that interfaces context representations with knowledge using a gated attention mechanism. We assess 
  the model's 
performance on 
  a recently published
  dataset for categorizing human needs.
  Selectively integrating  knowledge paths boosts performance and establishes a new state-of-the-art.
  Our model offers interpretability through the learned attention map over commonsense knowledge paths. Human evaluation highlights the relevance of the encoded knowledge.
  
  \end{abstract}

\section{Introduction}

Sentiment analysis and emotion detection are 
essential tasks in human-computer interaction.  
Due to its broad practical applications, there has been rapid growth in the field of sentiment analysis \cite{zhang2018deep}.  Although 
state-of-the-art sentiment analysis can detect the polarity of text units \cite{hamilton2016inducing, socher2013recursive}, there has been limited work towards explaining the reasons for the expression of sentiment and
emotions 
in texts \cite{li2017reflections}. 
In our work, we aim to go beyond the detection of sentiment,
toward explaining sentiments. Such explanations can range from detecting overtly expressed explanations or reasons for sentiments towards specific aspects of, e.g., products or films, as in user reviews to the explanation of the underlying reasons for emotional reactions of characters in a narrative story.
The latter requires 
understanding of stories and modeling 
the mental state of characters. 
Recently, \citet{ding2018human} proposed to categorize affective events 
with categories based on human needs, to provide explanations of people's attitudes towards
such events. Given an 
expression 
such as \textit{I broke my leg}, they categorize the reason for the expressed negative sentiment as being related to a need concerning `health'.

In this paper we focus on the \textit{Modelling Naive Psychology of Characters in Simple Commonsense Stories}
dataset of \citet{rashkin2018modeling}, which contains annotations of a fully-specified chain of motivations and emotional reactions of characters for
a collection of narrative stories. The 
stories are annotated with 
labels from multiple theories of psychology \cite{reiss2004multifaceted,maslow1943theory,plutchik:1980} 
to provide explanations for the emotional reactions of characters.

Similar to \citet{ding2018human}, we hypothesize that 
emotional reactions (joy, trust, fear, etc.) of characters 
can be explained by (dis)satisfaction 
of their psychological needs. 
However, predicting categories of human needs that underlie the expression of sentiment is a difficult task for a computational model. 
It requires not only detecting
surface patterns from the text, 
but also requires commonsense knowledge about how a given situation may or may not satisfy specific human needs of a character. 
Such knowledge can be diverse and complex, and will typically 
be implicit in the text. 
In contrast, human readers can make use of relevant information 
from the story and associate it with their knowledge about human interaction, 
desires and human needs, and thus will be able to infer underlying reasons for emotions indicated in the text. 
In this work, we propose 
a computational model that aims to categorize human needs of story characters by integrating commonsense knowledge 
from ConceptNet \cite{speer2012representing}.
Our model aims to imitate human understanding of a story, by (i) learning to select relevant words from the text, (ii) extracting pieces of
knowledge from the commonsense 
inventory and (iii) associating them with human need categories put forth by psychological theories. 
Our assumption is that by integrating commonsense knowledge in our model
we will be able to overcome the lack of textual evidence in establishing relations between expressed emotions in specific situations and the inferable human needs of story characters. In order to 
provide such missing associations,
we leverage the graph structure of the knowledge source. Since these 
connections can be diverse and 
complex, we develop a novel approach to extract and rank
multi-hop relation paths from ConceptNet
using graph-based methods. 

Our contributions are: (i) We propose a novel
approach to extract and rank multi-hop relation paths from a commonsense knowledge resource using graph-based features and algorithms. (ii) We present an end-to-end model enhanced with attention and a gated knowledge integration component to 
predict human needs in a given context. To the best of our knowledge, our model is the first to advance commonsense knowledge for this task. 
(iii) We conduct experiments that demonstrate the effectiveness of the extracted knowledge paths and show significant performance improvements over the prior state-of-the-art. (iv) Our model provides interpretability in two ways: by selecting relevant words from the input text and by choosing relevant knowledge paths from the imported knowledge. In both cases, the degree of relevance is indicated via an attention map. (v) A small-scale human evaluation demonstrates that the extracted multi-hop knowledge paths are indeed relevant.
Our code is made publicly available.\footnote{\url{https://github.com/debjitpaul/Multi-Hop-Knowledge-Paths-Human-Needs}}

\section{Related Work}

\textbf{Sentiment Analysis and Beyond.} Starting with 
\citet{pang2002thumbs}, sentiment analysis and emotion detection has grown to a wide
research field. 
Researchers have investigated polarity classification, sentiment and emotion detection and classification \cite{tang2015document, yin2017document, li2017end} on various levels 
(tokens, phrases, sentences or documents), as well as structured prediction tasks such as the identification
of  holders and targets \cite{deng2015joint} or sentiment inference \cite{choi2016document}. 
Our work goes beyond the analysis of overtly expressed sentiment and aims at identifying goals, desires or needs underlying the expression of sentiment.
\citet{li2017reflections} argued that the goals of an opinion holder can be categorized by human needs. There has been work related to goals, desires, wish detection \cite{goldberg2009may, rahimtoroghi2017modelling}.
Most recently, \citet{ding2018human} propose to categorize affective events into physiological needs to explain people's motivations and desires. \citet{rashkin2018modeling} published a dataset for tracking emotional reactions and motivations of characters in stories. In this work, we use this
dataset to develop a knowledge-enhanced system that `explains' sentiment in terms of human needs.

\textbf{Integrating structured knowledge into neural NLU systems.} Neural models aimed at solving NLU tasks have been shown to profit from the integration of knowledge, using different methods:
\citet{xu2017incorporating} show that injecting loosely structured knowledge with a recall-gate mechanism is beneficial for conversation
 modeling; \citet{P18-1076} and \citet{weissenborn2017dynamic} propose integration of commonsense knowledge for reading comprehension: the former explicitly encode selected triples from ConceptNet using attention mechanisms, the latter enriches question and context embeddings by encoding triples as mapped statements extracted from ConceptNet.
Concurrently to our work, \citet{bauer2018commonsense} proposed a heuristic method to extract multi-hop paths from ConceptNet for a 
reading comprehension task. They construct paths starting from concepts appearing in the question to concepts appearing in the context,  aiming to emulate multi-hop reasoning. 
\citet{tamilselvam2017graph} use ConceptNet relations for aspect-based sentiment analysis.
Similar to our approach, \citet{bordes2014question} make use of knowledge bases to obtain longer paths
connecting entities appearing in questions to answers in a QA task. They also provide a richer representation of answers by building subgraphs of entities appearing in answers. In contrast, our work aims to provide 
information about missing links between sentiment words in a text and underlying human needs by extracting relevant multi-hop paths from structured knowledge bases.

\section{Selecting and Ranking Commonsense Knowledge to Predict Human Needs}
\begin{figure}[t]
    \includegraphics[scale=0.25]{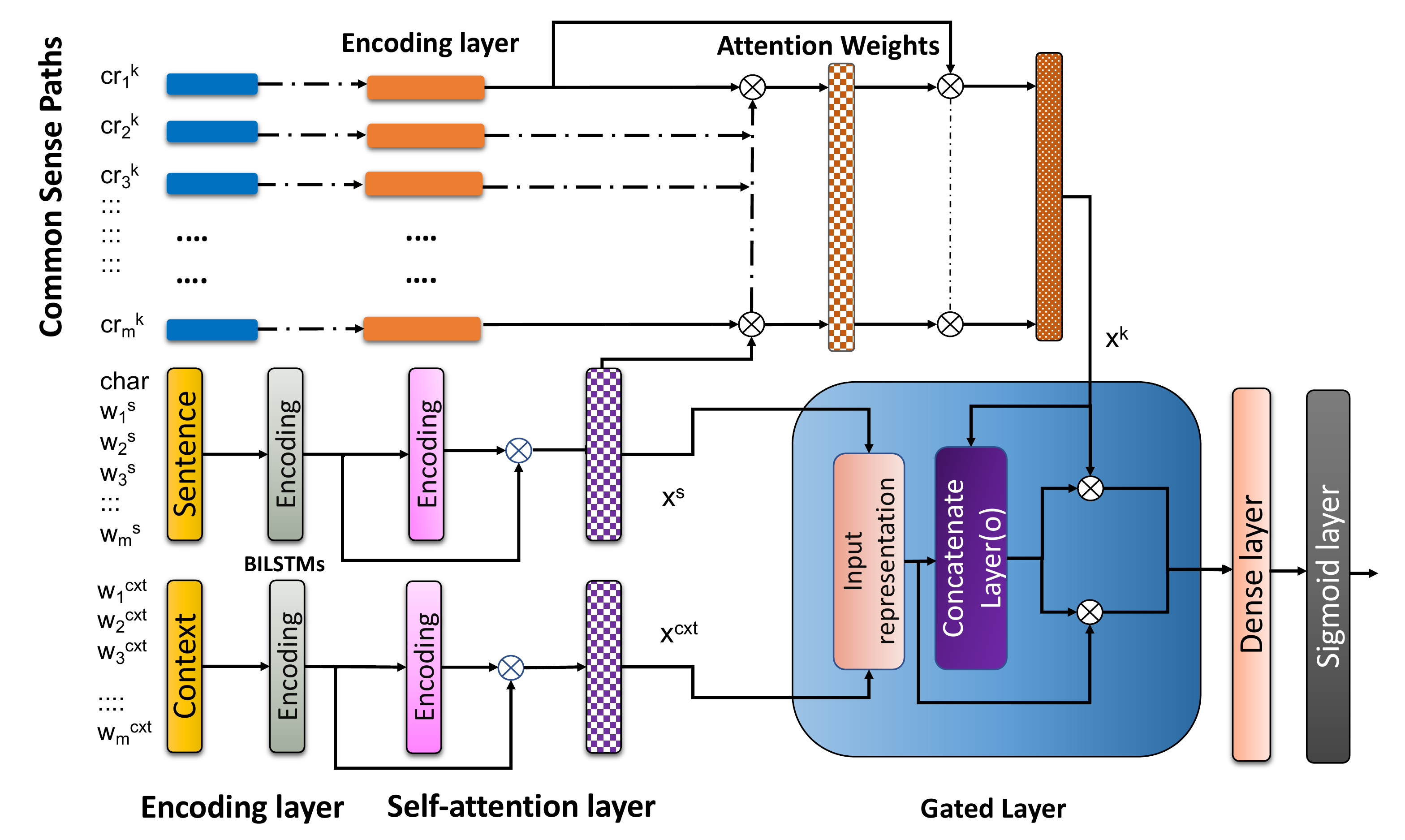}
    \caption{Attention over multi-hop knowledge paths.}
    \label{fig:model}
\end{figure}

Our task is to automatically predict
human needs of story characters 
given a story context.
In this task, following the setup of \citet{rashkin2018modeling}, 
we \textit{explain} the probable reasons for the expression of emotions 
by predicting appropriate categories from 
two 
theories of psychology:
\textit{Hierarchy of needs} \cite{maslow1943theory} and \textit{basic motives} \cite{reiss200216}. The task is defined as 
a multi-label classification problem
with five coarse-grained (Maslow) and 19 fine-grained (Reiss) 
categories, respectively (see Fig.\ \ref{fig:needs}).\footnote[1]{Details about the labels are given in the Supplement.} 
We 
start with a 
Bi-LSTM encoder with self-attention as a baseline model, to efficiently categorize human needs.  
We then show how to select and rank multi-hop commonsense knowledge paths from ConceptNet that connect textual expressions with human need categories. Finally, we
extend our model with a
gated knowledge integration mechanism to incorporate relevant multi-hop commonsense knowledge paths for predicting human needs. An overview of the model is given in Figure \ref{fig:model}. 
We now describe each component in detail.

\begin{figure}[t]
    \includegraphics[scale=0.35]{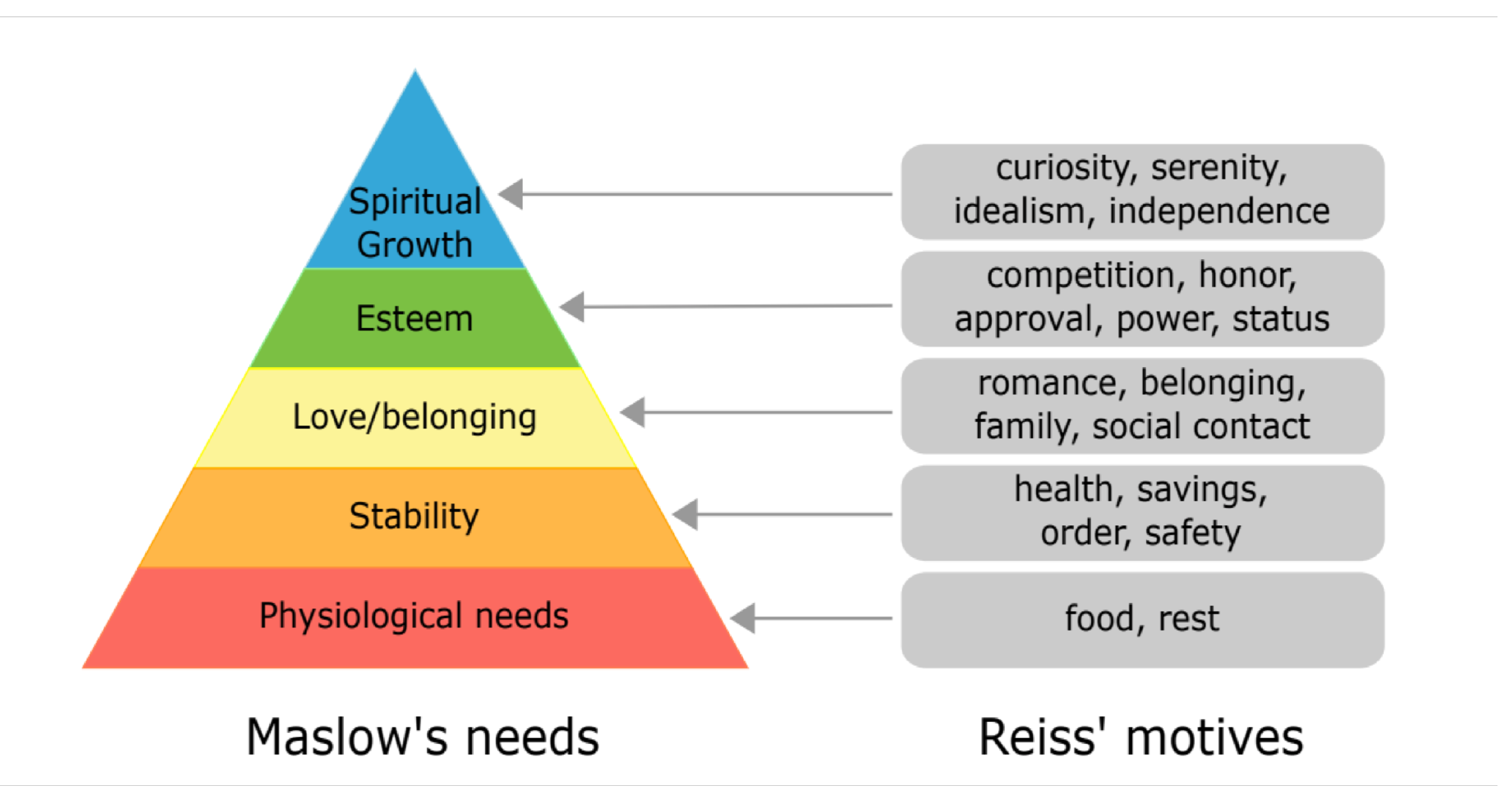}
    \caption{Maslow and Reiss: {Theories of Psychology} as presented in \citet{rashkin2018modeling}.}
    \label{fig:needs}
\end{figure}

\subsection{A Bi-LSTM Encoder with Attention to Predict Human Needs}
Our Bi-LSTM encoder takes as input a sentence $S$ consisting of a sequence of tokens, denoted as $w_{1}^s, w_{2}^s, ...., w_{n}^s$, or $w_{1:n}^s$ and its preceding context \(Cxt\), denoted as \(w_{1}^{cxt}, w_{2}^{cxt}, ...., w_{m}^{cxt}\), or $w_{1:m}^{cxt}$.
As further input we read the name of a story character, which is concatenated to the input sentence.
For this input the model is tasked to predict appropriate human need category labels ${z \in Z}$, according to a predefined inventory.\\
\if false
As discussed earlier, for a model to predict the human needs of a story character, it needs to be able to do the following : effectively read and encode each tokens in the text, focus on specific important tokens in the text, and ability to read and incorporate commonsense knowledge relations about the concepts appeared in the text and human needs to predict the needs of the character. 
\fi

\textbf{Embedding layer:}
We embed each word from the sentence and the context with a contextualized word representation using character-based word representations 
(ELMo) \cite{Peters:2018}. The embedding of each word $w_i$ in the sentence and context is represented as \(e^s_i\) and 
\(e^{cxt}_i\), respectively.

\textbf{Encoding Layer:}
We use a single-layer Bi-LSTM \cite{hochreiter1997long} to obtain 
sentence and context 
representations \(h^s\) and \(h^{cxt}\), which we
form by concatenating the final states of the forward and backward 
encoders. 
\begin{equation}
    h^s = BiLSTM(e^s_{1:n}); h^{cxt} = BiLSTM(e^{cxt}_{1:m})
\end{equation}

\textbf{A Self-Attention Layer}
allows the model to dynamically control how much each token contributes to the sentence and context representation. We use a modified version of self-attention proposed by \citet{rei2018zero}, where both input
representations are passed through a feedforward layer to generate scalar values for each word in context \(v^{cxt}_i\) and sentence  \(v^s_i\) (cf.\ (2-5)). 
\begin{equation}
     a^s_i = ReLU({W^s_i}{h^s_i}+b^s_i),
\end{equation}
\begin{equation}
     a^{cxt}_i = ReLU({W^{cxt}_i}{h^{cxt}_i}+b^{cxt}_i)
\end{equation}
\begin{equation}
    v^s_i=W^s_v{_i}a^s_i+ b^s_v{_i}
\end{equation}
\begin{equation}
     v^{cxt}_i=W^{cxt}_v{_i}a^{cxt}_i + b^{cxt}_v{_i}
\end{equation}

where, \( W^s,b^s, W^{cxt}, b^{cxt}, W^s_v, W^{cxt}_v  \) are trainable parameters. We calculate the soft attention weights for both sentence and context: 
\begin{equation}
    \label{eqn:sig}
    \widetilde{v_i} = \frac{1}{1+exp(-{v_i})}; ~~
    {\hat{v}_{i}}=\frac{\widetilde{v}_{i}}{\sum_{k=1}^N \widetilde{v}_{k}}
\end{equation}
where, \(\widetilde{v}_{i}\) is the output of the sigmoid function, therefore 
\(\widetilde{v}_{i}\) 
is in the range [0,1]
and \(\hat{v}_{i}\) is the normalized version of \(\widetilde{v}_{i}\). Values \(\hat{v}_{i}\) are used as attention weights to obtain the final
sentence and context representations \(x^s\) and \(x^{cxt}\), respectively: 
\begin{multicols}{2}
  \begin{equation}
    x^s = \sum_{i=1}^N{\hat{v_i}^sh^s_{i}}
  \end{equation}\break
  \begin{equation}
    x^{cxt} = \sum_{i=1}^M{\hat{v_i}^{cxt}h^{cxt}_{i}}
  \end{equation}
\end{multicols}
with \(N\) and \(M\) the number of tokens in $S$ and $Cxt$. The output of the self-attention layer is generated by concatenating $x^s$ and $x^{cxt}$. We pass this 
representation 
through a FF layer of dimension \(Z\): 
\begin{equation}
    y = ReLU(W_y[x^s;x^{cxt}]+b_y)
    \label{eqn:input_representation}
\end{equation}
where \(W_y, b_y\) are trainable parameters and ';' denotes concatenation of two vectors. Finally, we feed the output layer \(y\) to a logistic regression layer to predict a binary label for each class \(z\in Z \), where \(Z\) is the set of category labels for a particular psychological theory (Maslow/Reiss, Fig.\ \ref{fig:needs}). 
\subsection{Extracting Commonsense Knowledge}
To improve the prediction capacity of our model,  we aim to leverage external commonsense know\-ledge that connects 
expressions from the sentence and context to human need categories. 
For this purpose we extract 
%
multi-hop commonsense knowledge paths that connect words in the textual inputs 
with the offered 
human need categories, using as resource ConceptNet \cite{speer2012representing}, a large commonsense knowledge inventory.
Identifying 
contextually relevant information from such a large knowledge base is a non-trivial task. We propose an effective two-step method to extract multi-hop knowledge paths that associate
concepts from the text
with human need categories: 
(i) collect all potentially relevant knowledge relations among concepts and human needs in a subgraph for each input sentence; (ii) rank, filter and select high-quality paths using graph-based local measures and graph centrality algorithms. 
\subsubsection{Construction of Sub-graphs}
ConceptNet is a graph \(G = (V, E)\) whose nodes are concepts and edges are relations between concepts (e.g. \textsc{Causes, MotivatedBy}).
For each sentence \(S\) we induce a subgraph \(G' = (V', E')\) where $V'$ comprises 
all concepts \(c \in V\) that appear in $S$ and the 
directly preceding sentence in  context $Cxt$. $V'$ also includes all concepts $c \in V$ that correspond to one of the human need categories in our label set $Z$. Fig.\ \ref{fig:cps} shows an example.\\
The sub-graph is constructed as follows:

\textbf{Shortest paths:} In a first step, we find all shortest paths $p'$ from ConceptNet that connect any 
concept \(c_{i}\in{V'}\) to any other concept \(c_{j}\in{V'}\) and 
to each human needs concept \(z\in{Z}\). We further include in $V'$ 
all the concepts $c \in V$ which are contained in the above shortest paths $p'$.

\textbf{Neighbours:} To better represent the meaning of the concepts in $V'$, we further include in $V'$ all concepts $c \in V$ that are directly connected to any $c \in V'$ that is not already included in $V'$.


\textbf{Sub-graph:} We finally construct a connected sub-graph $G'=(V',E')$ from $V'$ by defining $E'$ as the set of all ConceptNet edges $e \in E$ that directly connect any pair of concepts $(c_i, c_j) \in V'$. 

Overall, we obtain a sub-graph that contains relations and concepts which are supposed to be useful to ``explain" 
\textit{why} and \textit{how strongly} 
concepts \(c_{i}\) that appear in the sentence and context are associated with any of the human needs \(z\in{Z}\).  

\subsubsection{Ranking and Selecting Multi-hop Paths}\label{selecting_know}
\begin{figure}[t]
  \centering
    \includegraphics[scale=0.37,height=6.2cm]{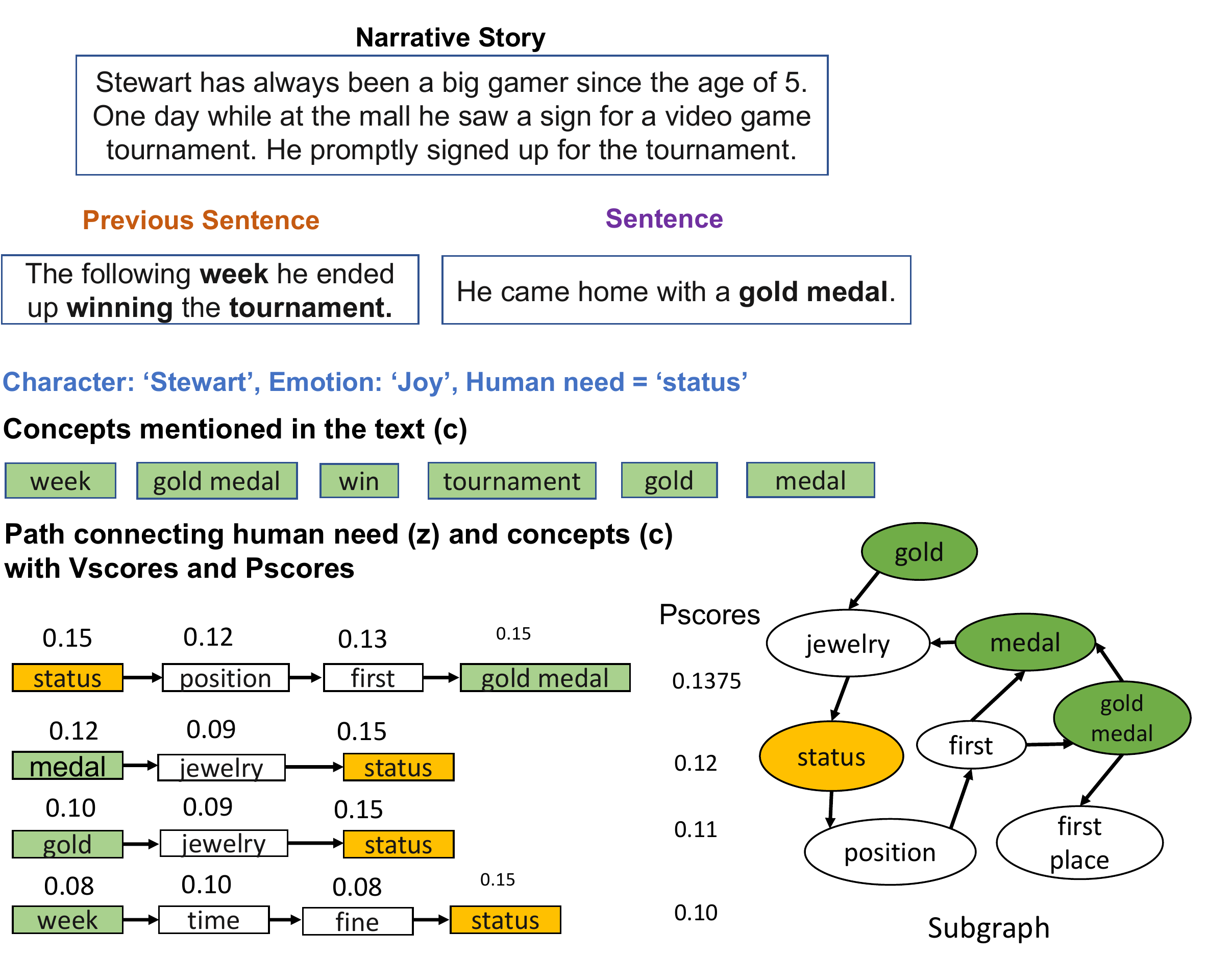}
    \caption{Illustration of
    commonsense path selection. Top: Context and sentence, Bottom: Selected knowledge paths with \textit{Vscores} and \textit{Pscores} (left) and the corresponding subgraph.
    Concepts from
    the text are marked green;
    yellow boxes show
    the human need label \textit{status} assigned to 
    \textit{Stewart}.}
    \label{fig:cps}
\end{figure}

We could use
all possible paths $p$ contained in the sub-graph $G'$, connecting concepts $c_i$ from the text and human needs concepts $z$ contained in $G'$,
as additional evidence to predict suitable human need categories.
But not all of them may be relevant. In order to select the most relevant paths, we propose a two-step method: (i) we score each vertex  
with a score (\textit{Vscore}) that reflects its importance in the sub-graph 
and on the basis of the vertices'  \textit{Vscores} we determine a path score \textit{Pscore}, as shown in Figure \ref{fig:cps}; (ii) we select the top-k paths with respect to the computed path score (\textit{Pscore}) . 

\textbf{(i) Vertex Scores and Path Scores:} 
We hypothesize that the
most useful commonsense relation paths should include 
vertices 
that are \textit{important} with respect to the entire extracted subgraph. 
We measure the importance of a vertex using 
different 
local graph measures: the \textit{closeness centrality measure, page rank} or \textit{personalized page rank}.

\textbf{Closeness Centrality (CC)} \cite{bavelas1950communication} reflects 
how close a vertex is to all other vertices in the given graph. It measures the average length of the shortest paths between a given vertex \(v_i\) and all other vertices in the given graph \(G'\).
In a connected graph, the closeness centrality \(CC(v_i)\) of a vertex \(v_i \in G'\) is computed as 
    \begin{equation}
         Vscore_{CC}(v_i) = \frac{\mid{V'}\mid}{\sum_j d\left(v_j,v_i\right)}
    \end{equation}
  where \({\mid{V'}\mid}\) represents the number of vertices in the graph \(G'\) and \(d(v_j,v_i)\) represents the length of the shortest path between \(v_i\) and \(v_j\). For each path we compute the normalized sum of \textit{Vscore}$_X$ of all vertices  \(v_j\) contained in the path, for any measure $X$ $\in \{ CC, PR, PPR\}$.
  \begin{equation}
   \label{eq:pathscore}
    Pscore_{X}= \frac{\sum_j Vscore_{X}(v_j)}{N}
  \end{equation}
  We rank the paths according to their \textit{Pscore}$_{CC}$,
  assuming
  that relevant paths will contain vertices that are close to the center of the sub-graph \(G'\). 
  
\textbf{PageRank (PR)}~\cite{brin1998anatomy} is a graph centrality algorithm that
measures the relative importance of a vertex in a 
graph. 
  The Page\-Rank score of a vertex \(v_i \in G'\) is computed as: 
    \begin{equation}
       Vscore_{PR}(v_i) = \alpha\sum_j{u_{ji}}\frac{v_j}{L_j}+\frac{1-\alpha}{n}
    \end{equation}
  where \(L_j =\sum_i u_{ji}\) is the number of neighbors of vertex $j$, \(\alpha\) is a damping factor representing the probability of jumping from a given vertex $v_i$ to another random vertex in the graph and $n$ represents the number of vertices in 
  \(G'\). We calculate \textit{Pscore}$_{PR}$ using Eq.\ \ref{eq:pathscore} and order the paths according to their \textit{Pscore}$_{PR}$,
  assuming
  that relevant paths will contain vertices with high relevance, as reflected by a high number of incoming edges.
  
 \textbf{Personalized PageRank (PPR)}~ \cite{haveliwala2002topic} 
 is used to determine the importance of a vertex with respect to a certain topic (set of vertices). Instead of assigning equal probability for a random jump \(\frac{1-\alpha}{n}\), PPR assigns stronger probability to certain vertices to prefer
 topical vertices.  The 
 PPR score of a vertex \(v \in G'\) is computed as: 
     \begin{equation}
        Vscore_{PPR}(v_i) = \alpha\sum_j{u_{ji}}\frac{v_j}{L_j}+\left({1-\alpha}\right){T}
     \end{equation}
  where \(T = \frac{1}{\mid{T_j}\mid}\) if nodes \(v_i\) belongs to topic ${T_j}$ and otherwise \(T =0\). 
  In our setting, $T_j$ will contain concepts from the text and human needs, to assign them higher probabilities.
  We calculate \textit{Pscore$_{PPR}$}
  using Eq.\ \ref{eq:pathscore} and order the paths according to their scores, assuming that relevant paths should contain vertices holding importance with respect to vertices representing
  concepts from the text and human needs. 
 
 \textbf{(ii) Path Selection:}
  We rank knowledge paths based on their \textit{Pscore} using the above relevance measures, and construct ranked lists of paths of two types: (i) paths connecting a human needs concept $z \in Z$ to a concept mentioned in the text ($p_{c-z}$) \footnote[2]{$p_{c-z}$ denotes path connecting a human needs concept $z \in Z$ and a concept \(c\) mentioned in the text.} and (ii) paths connecting concepts in the text ($p_{c-c}$) \footnote[3]{$p_{c-c}$ denotes path connecting a concept \(c\) and another concept \(c\) mentioned in the text.}. 
  Ranked lists of paths are constructed individually for concepts that constitute the start or endpoint of a path: a human needs concept for $p_{c-z}$ or any concept from the text for $p_{c-c}$. 

Figure \ref{fig:cps} 
  illustrates an example where the character \textit{Stewart} felt \textit{joy} after winning a gold medal. The annotated human need label is \textit{status}. We show the paths selected by our algorithm that connect concepts from the text and the human need \textit{status}. We select the top-$k$ paths of type $p_{c-z}$
  for each human need to
  capture relevant knowledge about human needs in relation to
  concepts in the text. 
  Similarly, we select the top-$k$ paths of type $p_{c-c}$ 
  for each $c_i$ to capture 
  relevant
  knowledge about the 
  text (not shown in Fig.\ 3).
\subsection{Extending the Model with Knowledge}
We have seen how to obtain a ranked list of commonsense knowledge paths from a subgraph  extracted from ConceptNet that connect concepts from the textual input and possible human needs categories that are the system's classification targets. Our intuition is that the extracted commonsense knowledge paths will provide useful evidence for our model to 
\textit{link the content expressed in the text to appropriate human need categories}. Paths that are selected by the model as a relevant connection between the input text and the labeled human needs concept can thus provide \textit{explanations} for emotions or goals expressed in the text \textit{in view of a human needs category}. 
We thus integrate these knowledge paths into our model,
(i) to help the model making correct predictions and (ii) to provide explanations of emotions expressed in the text in view of different human needs categories.
For each input, we represent the extracted ranked list of $n$ commonsense knowledge paths $p$ as a list \({cr^{k,1}, cr^{k,2}, ...., cr^{k,n}}\), where
each \(cr^{k,i}_{1:l}\) 
represents a path consisting of concepts and relations,
with $l$ the length of the path. We embed all concepts and relations in  \(cr^{k,i}_{1:l}\)  
with pretrained GloVe \cite{pennington2014glove} embeddings.

\textbf{Encoding Layer:}
We use a 
single-layer BiLSTM
to obtain 
encodings (\(h^{k,i}\)) for each knowledge path 
\begin{equation}
    h^{k,i} = BiLSTM(e^{k,i}_{1:n})
\end{equation}
where \(h^k\) represents the output of the BiLSTM for the knowledge path and $i$ its
the ranking index.

\textbf{Attention layer:}
We use an attention layer, where each encoded commonsense knowledge path interacts with the sentence representation \(x^s\) to receive attention weights (\(\hat{h}^{k,i}\)): 
\begin{equation}
    \label{eqn:sigmoid}
    \widetilde{h}^{k,i} = \sigma(x^s h^{k,i}),~~~~ \hat{h}^{k,i} = \frac{\widetilde{h}^{k,i}}{\sum_{i=1}^N \widetilde{h}^{k,i}}
\end{equation}
In Eq.\ \ref{eqn:sigmoid}, we use sigmoid to calculate the attention weights, similar to Eq.\ \ref{eqn:sig}. However, this time we compute attention to highlight which knowledge paths are important for a given input representation ($x^s$ being the final state hidden representation over the input sentence, Eq. 7).
To obtain the sentence-aware commonsense knowledge representation \(x^k\), we pass the output of the attention layer through a feedforward layer. \(W_{k}\), \(b_k\) are trainable parameters. 
\begin{equation}
    x^k = ReLU(W_k (\sum_{i=1}^N{\hat{h}^{k,i}h^{k,i}) + b_k)}
\end{equation}
\subsection{Distilling knowledge into the model}
In order to incorporate the selected and weighted knowledge into the model, 
we concatenate the sentence \(x^s\), context \(x^{cxt}\) and knowledge \(x^k\) representation and pass it through a FF layer.
\begin{equation}
    o_i = {ReLU(W_z[x_i^s;x_i^{cxt};x_i^k]+b_z)}
\end{equation}
We employ a gating mechanism 
to allow the model to selectively incorporate relevant information from commonsense knowledge $x^k$ and from the joint input representation $y_i$ (see Eq. \ref{eqn:input_representation}) \textit{separately}.
We finally pass it to a logistic regression classifier to predict a binary label for each class \(z\) in the 
set \(Z\) of category labels 

\begin{equation}
    z_i = \sigma(W_{\widetilde{y}_z}(o_i \odot y_i + o_i \odot x^k_i)+b_{\widetilde{y}_z})
\end{equation}
where \(\odot\) represents element-wise multiplication, \(b_{\widetilde{y}_z}\) , \(W_{\widetilde{y}_z}\) are trainable parameters.

\section{Experimental Setup}
\begin{table}[t!]
\centering
\small
\scalebox{0.9}{
\begin{tabular}{@{}llll@{}}
\toprule
{\bf Classification} & {\bf Train}& {\bf Dev} & {\bf Test}\\
\midrule
Reiss & {5432}& {1469}& {5368} \\
Reiss without \textit{belonging} class & {5431}& {1469}& {5366} \\
Maslow & {6873}&  {1882} & {6821} \\
\bottomrule
\end{tabular}}
\caption{Dataset Statistics:
nb.\ of instances 
(sentences with annotated characters and human need labels).}\label{tab:data_stat}
\end{table}
\textbf{Dataset:} We evaluate our model on the 
\textit{Modeling Naive Psychology of Characters in Simple Commonsense Stories (MNPCSCS)} dataset \cite{rashkin2018modeling}. It contains narrative stories where each sentence is annotated with a character and a set of human need categories from two inventories: Maslow's (with five coarse-grained) and Reiss's (with 19 fine-grained) categories
(Reiss's labels are considered as sub-categories of Maslow's).
The data contains the original worker annotations. Following
prior work we select the annotations that display the
``majority label" i.e., categories voted on by \(\geq 2\) workers.
Since no training data is available, similar to prior work we use a portion of the devset as training data, by 
performing a random split, using 
80\% 
of the data to train the classifier,
 and 20\% 
 to tune 
 parameters. Data statistics is reported in Table \ref{tab:data_stat}.

\citet{rashkin2018modeling} report that there is low annotator agreement i.a.\ between the \textit{belonging} and the \textit{approval} class. We also find
high co-occurrence of the \textit{belonging, approval} and \textit{social contact} classes, where \textit{belonging} and \textit{social contact} both pertain to the Maslow class \textit{Love/belonging} while \textit{approval} belongs to the Maslow class \textit{Esteem}. This indicates that \textit{belonging} interacts with \textit{Love/belonging} and \textit{Esteem} in relation to social contact.
We further observed during our study 
that in the Reiss dataset the number of instances annotated with the
 \textit{belonging} class 
 is very low (no.\ of instances in training is 24, and in dev 5). 
 The performance for this class is thus severely hampered, with 4.7 $F_1$ score for BiLSTM+Self-Attention and 7.1 $F_1$ score for BiLSTM+Self-Attention+Knowledge.
 After establishing benchmark results with prior work (cf.\ Table \ref{tab:all}, including \textit{belonging}), we perform all further experiments with a reduced Reiss dataset, by elimi\-na\-ting the \textit{belonging} class from all instances. This impacts the overall number of instances only slightly:
by \textit{one} instance for training and \textit{two} instances for test, as shown in Table \ref{tab:data_stat}.
\textbf{Training:}
During training we minimize the weighted binary cross entropy loss,
\begin{equation}
    L = \sum_{z=1}^Z{w_{z}y_{z}{log}\widetilde{y}_{z}+(1-w_{z})(1-y_{z})log(1-\widetilde{y}_{z})}
\end{equation}
\begin{equation}
    w_{z}=\frac{1}{1-exp^{-\sqrt{P(y_{z})}}}
\end{equation}
where $Z$ is the number of class labels in 
the 
classification tasks and $w_{z}$ is the weight. $P(y_{z})$ is the marginal class probability of a positive label for $z$ in the training set. 

\textbf{Embeddings:} To compare our model with prior work we experiment with pretrained GloVe (100d) embeddings \cite{pennington2014glove}. Otherwise we used GloVe (300d) and pretrained ELMo embeddings \cite{Peters:2018} to train our model.

\textbf{Hyperparameters for knowledge inclusion:} We compute ranked lists of knowledge paths of two types: $p_{c-z}$ and $p_{c-c}$.
We use the top-3 $p_{c-z}$ paths for each $z$ using our best ranking strategy (Closeness Centrality
+ 
Personalized PageRank) in our best system results (Tables \ref{tab:all}, \ref{tab:model_ab}, \ref{tab:first}), and also considered paths $p_{c-c}$ (top-3 per pair) when evaluating different path selection strategies (Table \ref{tab:ckp}).

\textbf{Evaluation Metrics:} We predict a binary label for each class using a binary classifier so the prediction of each label is conditionally independent of the other classes given a context representation of the sentence. In all
prediction tasks we report the micro-averaged Precision (P), Recall (R) and $F_1$ scores by counting the number of positive instances across all of the categories. All reported results
are averaged over five runs. More information on the dataset, metrics and all other training details are given in the Supplement.

\section{Results}
Our experiment results 
are summarized in Table \ref{tab:all}. 
We benchmark our baseline BiLSTM+Self-Attention model (BM, BM w/ knowledge) against the models proposed in \citet{rashkin2018modeling}: a BiLSTM and a CNN model, and models based on the  recurrent entity network (REN) \citep{DBLP:journals/corr/HenaffWSBL16} and neural process networks (NPN) \citep{bosselut2018published}.
The latter
differ from the basic encoding models (BiLSTM, CNN) and our own models by explicitly modeling entities.
We find that our baseline model BM outperforms all prior work, achieving new state-of-the-art results. 
For Maslow we show improvement of {21.02 pp.\ $F_1$ score}.
For BM+K this yields  
a boost of 6.39 and 3.15 pp.\ $F_1$ score for Reiss and Maslow, respectively. When using ELMo with BM we see an improvement in recall. However, adding knowledge on top improves the precision by 2.24 and 4.04 pp.\ for Reiss and Maslow.
In all cases, injecting knowledge
improves the model's precision and $F_1$ score.

\begin{table}[t!]
\small
\centering
\scalebox{0.89}{
\begin{tabular}{@{}l@{~}c@{~~}c@{~~} c@{~~} l@{~~} c@{~~} c@{~~} l@{}}
\toprule
& & \multicolumn{3}{@{}c}{Reiss} & \multicolumn{3}{c}{Maslow} \\
{\bf Model} &{\bf WE}&{\bf P}& {\bf R} & {\bf F1}&{\bf P}& {\bf R} & {\bf F1 } \\\midrule
BiLSTM$^\diamond$  &{G$_{\rm 100d}$} &{18.35}& {27.61}& {22.05} & {31.29}& {33.85}& {32.52} \\
CNN$^\diamond$  &{G$_{\rm 100d}$} &{18.89} & {31.22} & {23.54} & {27.47}& {41.01}& {32.09} \\
REN$^\diamond$  &{G$_{\rm 100d}$}& {16.79} &  {22.20} & {19.12} & {26.24}& {42.14}& {32.34} \\
NPN$^\diamond$  &{G$_{\rm 100d}$}&{13.13} & {26.44} & {17.55} & {24.27}& {44.16}& {31.33} \\
BM &{G$_{\rm 100d}$} &{25.08}& {28.25}& {26.57} &{47.65}& {60.98}& {53.54} \\
BM + K{$^\clubsuit$} &{G$_{\rm 100d}$}&{\bf 28.47}& {\bf 39.13}& {\bf 32.96} &{\bf 50.54}& {\bf 64.54}& {5\bf 6.69}\\
BM &{ELMo} &{29.50}& {\bf 44.28}& {35.41$_{\pm 0.23}$}& {53.86}& {\bf 67.23}& {59.81$_{\pm 0.23}$}\\
BM + K$^\clubsuit$ &{ELMo}&{\bf 31.74}& { 43.51}& {\bf 36.70$_{\pm 0.14}$} &{\bf 57.90}& {66.07}& {\bf 61.72$_{\pm 0.11}$}\\\hline
BM$^\star$ &{ELMo} &{31.45}& {44.29}& {37.70}\\
BM + K{$^\star$}{$^\clubsuit$} &{ELMo} &{36.76}& {42.53}& {39.44} \\
\bottomrule
\end{tabular}}
\caption{Multi-label Classification Results:
$^\diamond$: results in Rashkin et al.; $^\star$: 
w/o \textit{belonging};
BM: BiLSTM+Self-Att.; +K:w/ knowledge, $^\clubsuit$:ranking method CC+PPR. }\label{tab:all}
\end{table}
Table \ref{tab:all} (bottom) 
presents results for 
the reduced dataset,
after eliminating Reiss' label \textit{belonging}. 
Since 
\textit{belonging} is a rare class, we observe further improvements. We see the same trend: adding knowledge improves the precision of the model.    


\subsection{Model Ablations} 
To obtain better insight into the contributions of individual components of our models, we perform an ablation study (Table \ref{tab:model_ab}). Here and in all later experiments we use richer (300d) GloVe embeddings and the dataset w/o \textit{belonging}. We show results including and not including self-attention and knowledge components.
We find
that using self-attention over sentences and contexts is highly effective, which indicates that learning how much each token contributes helps the model to improve performance.
We observe that integrating knowledge improves the overall $F_1$ score and yields a gain in precision with ELMo. Further, integrating knowledge using the gating mechanism we see a considerable 
increase of 3.58 and 1.74 pp.\ $F_1$ score 
improvement over our baseline model for GloVe and ELMo representations respectively. 



\begin{table}[t!]
\small
\centering
\scalebox{0.95}{
\begin{tabular}{@{}llllccc@{}}
\toprule
{\bf WE} & {\bf Atten} &{\bf K}&{\bf Gated}&{\bf P}& {\bf R} & {\bf F1}\\
\midrule
G$_{\rm 300d}$ &-&-&-& {23.31}& {34.69}& {27.89} \\
G$_{\rm 300d}$ &\checkmark&-&-& {26.09}& {35.59}& {30.11} \\
G$_{\rm 300d}$ &\checkmark&\checkmark&-& {27.99}& {37.73}& {32.14} \\
G$_{\rm 300d}$ &\checkmark&\checkmark&\checkmark& \textbf{28.65}& \textbf{39.42}& \textbf{33.19} \\
ELMo &-&-&-& {32.35}& {42.66}& {36.80} \\
ELMo &\checkmark&-&-& {31.45}& {44.29}& {37.70} \\
ELMo &\checkmark&\checkmark&-& {32.65}& \textbf{45.60}& {38.05}
\\
ELMo &\checkmark&\checkmark&\checkmark& \textbf{36.76}& {42.53}& \textbf{39.44}  
\\\bottomrule
\end{tabular}}
\caption{Model ablations for Reiss Classification on \textit{MNPCSCS} dataset w/o \textit{belonging}.}\label{tab:model_ab}
\end{table}
\begin{table}[t!]
\centering
\small
\scalebox{0.95}{
\begin{tabular}{llccc}
\toprule
{\bf Path}&{\bf Ranking} & {\bf P}& {\bf R} & {\bf F1}\\
\midrule
S+M($P_{c-z}$+ $P_{c-c}$)&None& {32.51}& {42.70}& {36.90} \\
S+M($P_{c-z}$+ $P_{c-c}$)&Random &{31.63} & {43.35} & {36.57} \\
\midrule
Single Hop($P_{c-z}$) &CC + PPR &{33.00}& {44.63}& {37.94} \\
S+M($P_{c-c}$  + $P_{c-z}$) &CC + PPR & {35.30}&{44.11}&{39.21}\\
\midrule
S+M($P_{c-z}$) &CC &{33.45}& \bf{47.93}& {39.40} \\
S+M($P_{c-z}$) &PR &{35.51}& {42.82}& {38.82} \\
S+M($P_{c-z}$) &PPR &{36.23}&{43.09}&{39.34}\\
S+M($P_{c-z}$) &CC + PPR &\bf{36.76}& {42.53}& \bf{39.44}\\
\bottomrule
\end{tabular}}
\caption{Results for different path selection strategies on \textit{MNPCSCS}
w/o \textit{belonging}; S+M:Single+Multi hop.}\label{tab:ckp}
\end{table}

\subsection{Commonsense path selection}
We further examine model performance for (i) different variants of selecting commonsense knowledge, including (ii) the effectiveness of the 
relevance ranking strategies discussed in \textsection \ref{selecting_know}. In Table \ref{tab:ckp},
rows 3-4 use our best ranking method: CC+PPR; rows 5-8 show results when using the top-3 ranked $p_{c-z}$ paths for each human need $z$ with different ranking measures.
\textit{None}
 shows results when no selection is applied to the set of extracted knowledge paths (i.e., using all possible paths from $p_{c-z}$ and $p_{c-c}$).
\textit{Random} randomly selects 3 paths for each human need from the set of paths used in \textit{None}. This yields only a slight drop in performance.
This 
suggests that not every path is relevant.
We evaluate the performance when only considering single-hop paths (now top-3 ranked using CC+PPR) 
 \textit{(Single-Hop)}. 
 We see an improvement over random paths and no selection, but not important
 enough. 
 In contrast, using both single and multi-hop paths in conjunction with relevance ranking improves the performance considerably
 (rows 4-8). 
This demonstrates that multi-hop paths are informative.
We also experimented with $p_{c-c}$+$p_{c-z}$. We find improvement in recall, however the overall performance decreases by 0.2 $F_1$ score compared to paths $p_{c-z}$ ranked using CC + PPR. 
Among different ranking measures \textit{precision} for Personalized PageRank performs best in comparison with CC and PR in isolation, and recall for CC in isolation is highest. 
Combining
CC
and PPR 
yields the best results among the different ranking strategies (rows 5-8). 



\section{Analysis}
\subsection{Performance per human need categories}
\begin{figure}[t]
    \includegraphics[width=7.6cm]{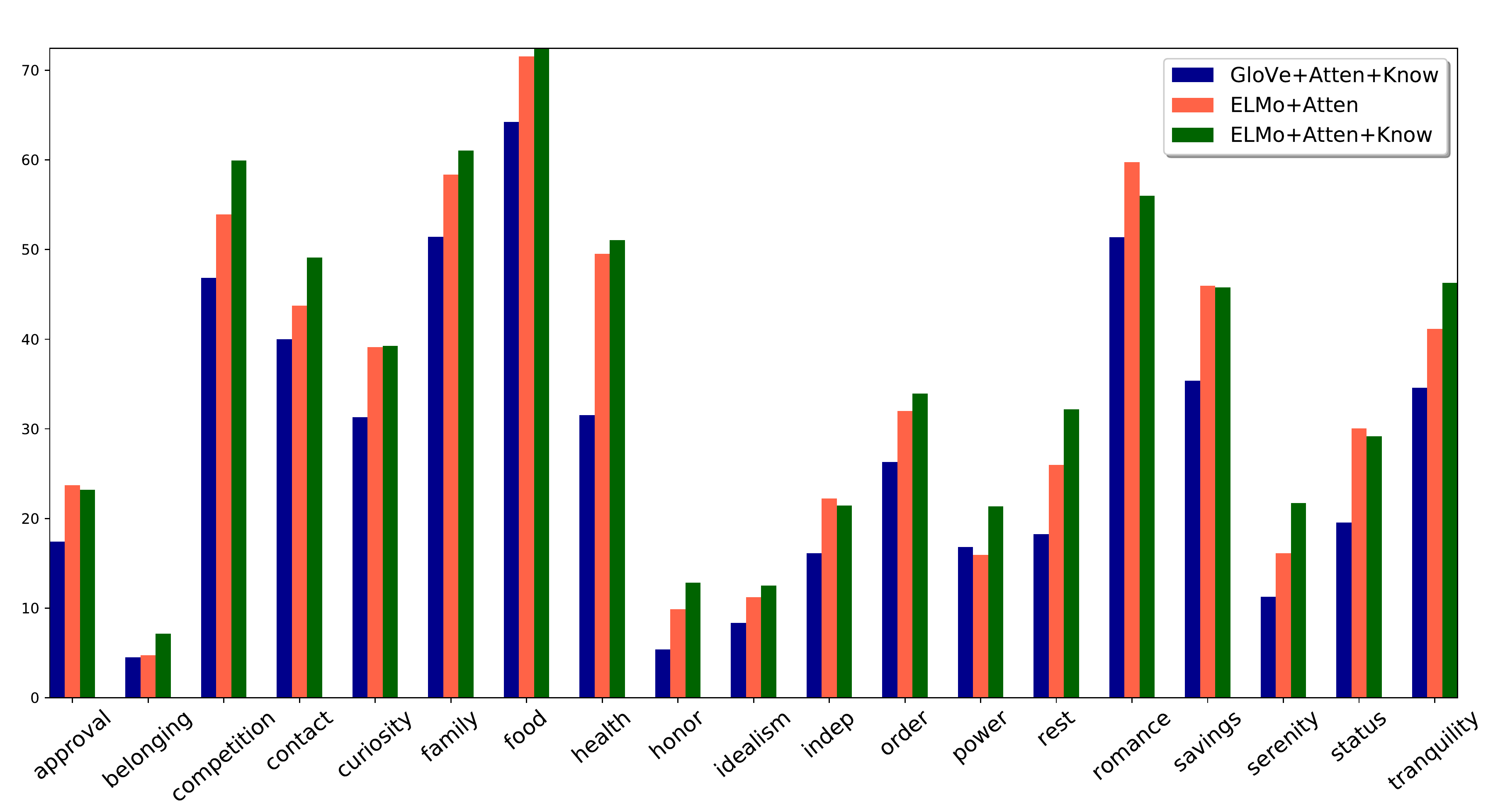}
    \caption{Best model's performance per human needs (F\(_1\) scores) for Reiss on \textit{MNPCSCS} dataset.}
    \label{fig:phn}
\end{figure}

We 
examined the model performance on each category 
(cf.\ Figure \ref{fig:phn}).
The model performs well 
for basic needs like \textit{food}, \textit{safety}, \textit{health}, \textit{romance}, etc. 
We note that inclusion of knowledge improves the performance for most classes (only 5 classes do not profit from knowledge compared to only using ELMo), especially for labels which are rare like \textit{honor, idealism, power.} We also found that the annotated labels can be subjective. 
For instance, \textit{Tom lost his job} is annotated with \textit{order} while our model predicts
\textit{savings}, which we consider to be correct.
Similar to
\citet{rashkin2018modeling} we
observe 
that preceding context helps the model to better predict
the characters' needs, e.g., \textit{Context: Erica's 
[..]
class had a reading challenge [..].
If she was able to read 50 books [..]
she won a pizza party!; Sentence: She read a book every day for the entire semester} is annotated with \textit{competition}. Without context the predicted label is \textit{curiosity}, however when including context, the model predicts \textit{competition, curiosity}. 
We measure the model’s performance when applying it only to the first sentence of each story (i.e., without the context). As shown in Table \ref{tab:first}, also in this setting the inclusion of knowledge improves the performance.

\begin{table}[h]
\small
\centering
\scalebox{0.9}{
\begin{tabular}{@{}llccc@{}}
\toprule
{\bf Model} & {\bf WE} &{\bf P}& {\bf R} & {\bf F1}\\
\midrule
{BM} &{ELMo}& {33.39}& {45.15} & {38.39}\\
{BM+K} &{ELMo}& {36.36}& {44.02} & {39.83}\\
\bottomrule
\end{tabular}}
\caption{Multi-label classification 
on \textit{MNPCSCS} 
w/o \textit{belonging class} and w/o context (1$^{st}$ sentence only)}. \label{tab:first}
\end{table}

\begin{figure}[t]
 \centering
\minipage{0.50\textwidth}
  \scriptsize{\textbf{Context:} Timmy had to renew his driver's license. He went to his local DMV. He waited in line for nearly 2 hours. He took a new picture for his driver's license.} \\\scriptsize{\textbf{Sentence:} He drove back home after an exhausting day.}\\
  \scriptsize{\textbf{True Label:} \textit{rest}}\\
  \scriptsize{\textbf{Predicted Label (BM):} \textit{status, approval, order}}\\
  \scriptsize{\textbf{Predicted Label (BM+K):} \textit{rest}}
\endminipage\hfill
\vspace*{-2mm}
\minipage{0.2\textwidth}
  \centering
  \includegraphics[width=8.0cm]{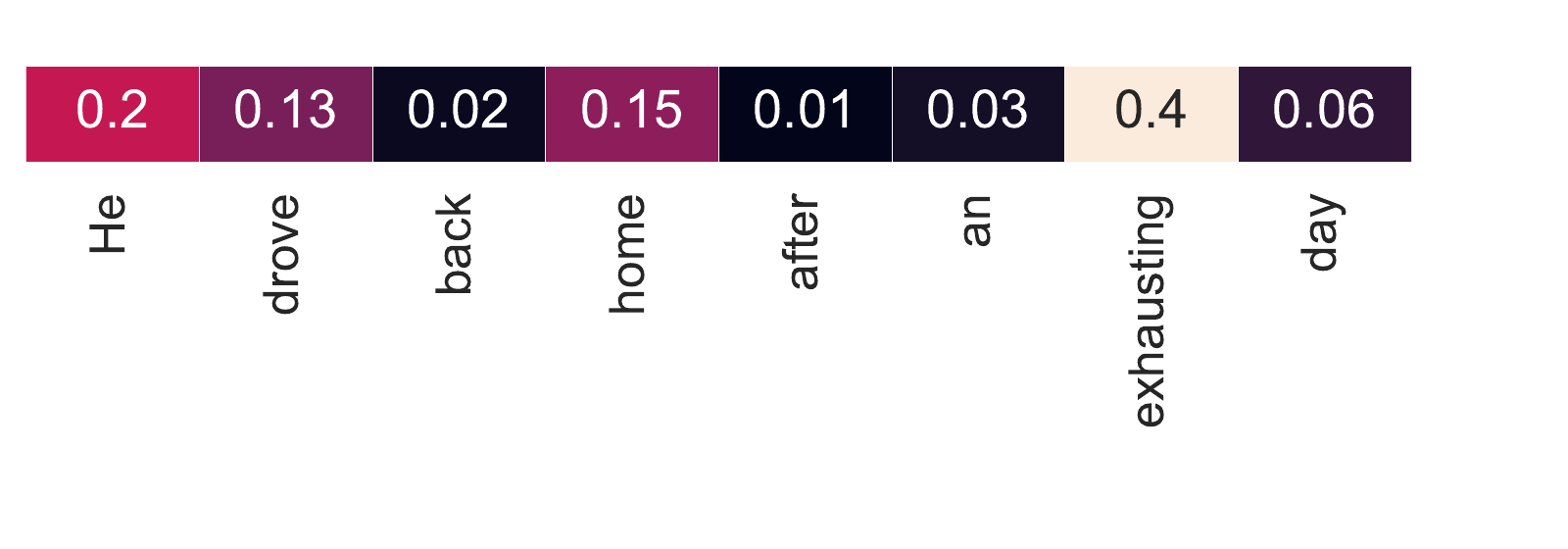}
\endminipage\hfill
\vspace*{-2mm}
\minipage{0.5\textwidth}
  \centering
  \includegraphics[width=\linewidth,height=4.0cm]{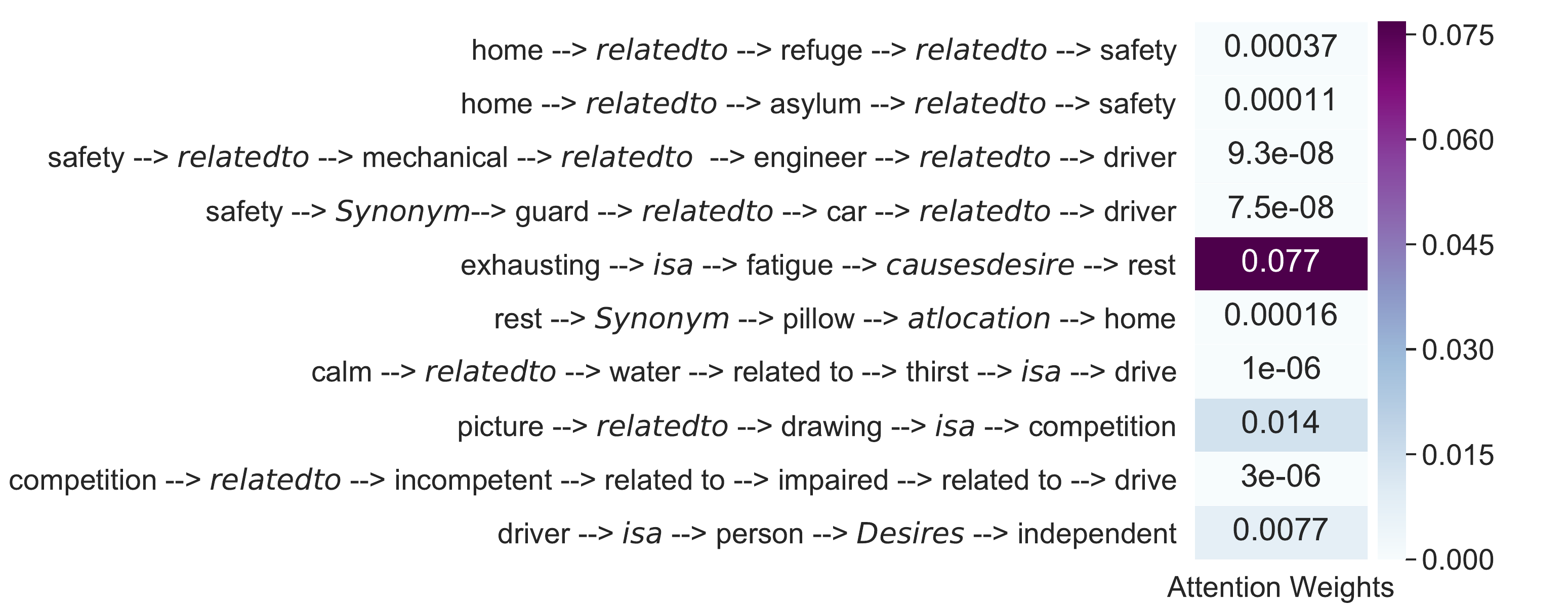}
\endminipage\hfill
\caption{Interpreting the attention weights on sentence representation and selected commonsense paths.}
\label{fig:interpretation_1}
\end{figure}

\subsection{Human evaluation of 
extracted paths}
We conduct human evaluation to test the effectiveness and relevance of the extracted commonsense knowledge paths. 
We randomly selected 50 
sentence-context pairs with their gold labels from the devset and 
extracted knowledge paths that contain the gold label (using 
CC+PPR for ranking). 
We asked three expert 
evaluators 
to decide whether the 
paths are relevant to provide information about the missing links between the concepts  in the sentence and the human need (gold label). The
inter-annotator agreement had a Fleiss' \(\kappa\)= 0.76.
The result for this evaluation shows that in 34\% of the cases computed on the basis of majority agreement, our algorithm was able to select a relevant commonsense path. More details about the human evaluation are given in the Supplement.

\subsection{Interpretabilty}
Finally we study the learned attention distributions of the interactions between sentence representation and knowledge paths,
in order to interpret how knowledge
is employed to make predictions.
Visualization of the
attention maps gives evidence of the ability of the model to capture relevant knowledge that connects  human needs to the input text. 
The model provides interpretability in two ways: by selecting tokens from the input text using Eq.\ref{eqn:sig} and by choosing knowledge paths from the imported knowledge using Eq.\ref{eqn:sigmoid} as shown in Figure \ref{fig:interpretation_1}. Figure \ref{fig:interpretation_1} shows an example where including knowledge paths helped the model to predict the correct human need category. The attention map depicts which exact paths are selected to make the prediction. In this example, the model correctly picks up the token \textit{``exhausting''} from the input sentence and the knowledge path \textit{``exhausting is a fatigue causes desire rest''}. 
We present more examples of extracted knowledge and its attention visualization in the Supplement.

\section{Conclusion}
We have introduced an effective new method to rank multi-hop relation paths from a commonsense knowledge resource using graph-based 
algorithms. 
Our end-to-end model
incorporates multi-hop knowledge paths to predict human needs. 
Due to the attention mechanism we can analyze the 
knowledge paths that the model considers in prediction. This enhances transparency and interpretability of the model.
We provide quantitative and qualitative evidence of the effectiveness of the extracted knowledge paths. We believe our relevance ranking strategy to select multi-hop knowledge paths can be beneficial for other NLU tasks. In future work, we will investigate structured and unstructured knowledge
sources to find explanations for sentiments and emotions.

\paragraph{Acknowledgements}
This work has been supported by the German Research Foundation as part of the Research
Training Group “Adaptive Preparation of Information from Heterogeneous Sources” (AIPHES)
under grant No.\ GRK 1994/1. We thank NVIDIA Corporation for donating GPUs used in this research. We thank \'{E}va M\'{u}jdricza-Maydt, Esther van den Berg and Angel Daza for evaluating the paths, and Todor Mihaylov for his valuable feedback throughout this work.


\bibliography{refs}
\bibliographystyle{acl_natbib}
\appendix
\section{Supplement Material}
\begin{figure*}[ht!]
  \centering
    \includegraphics[scale=0.5]{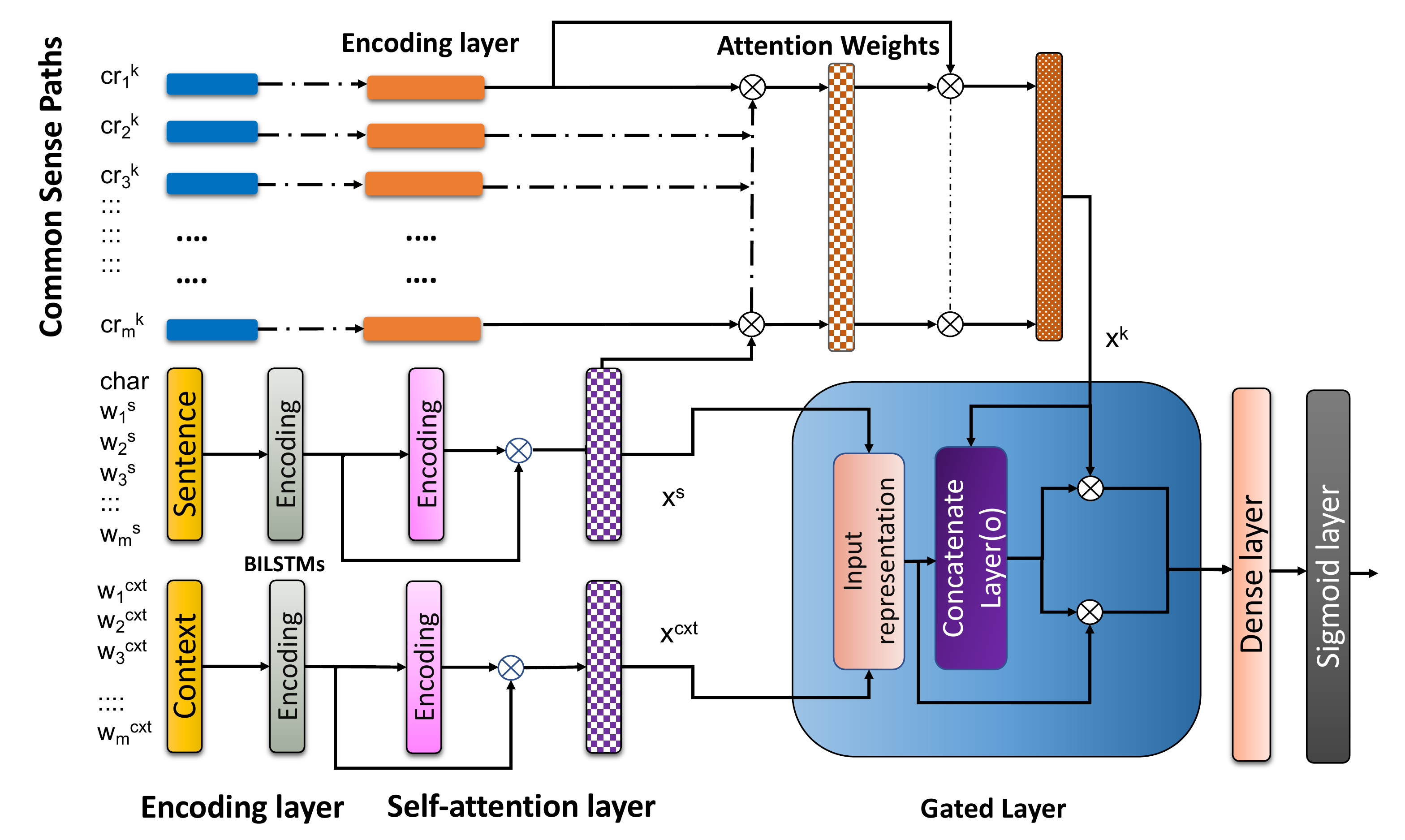}
    \caption{Full model}
    \label{fig:full_model}
\end{figure*}

A detailed visualization of our model, described in
Section 3 of the main paper is shown in Fig. \ref{fig:full_model}.

\subsection{Dataset Details}
\begin{figure}[ht!]
  \centering
    \includegraphics[scale=0.35]{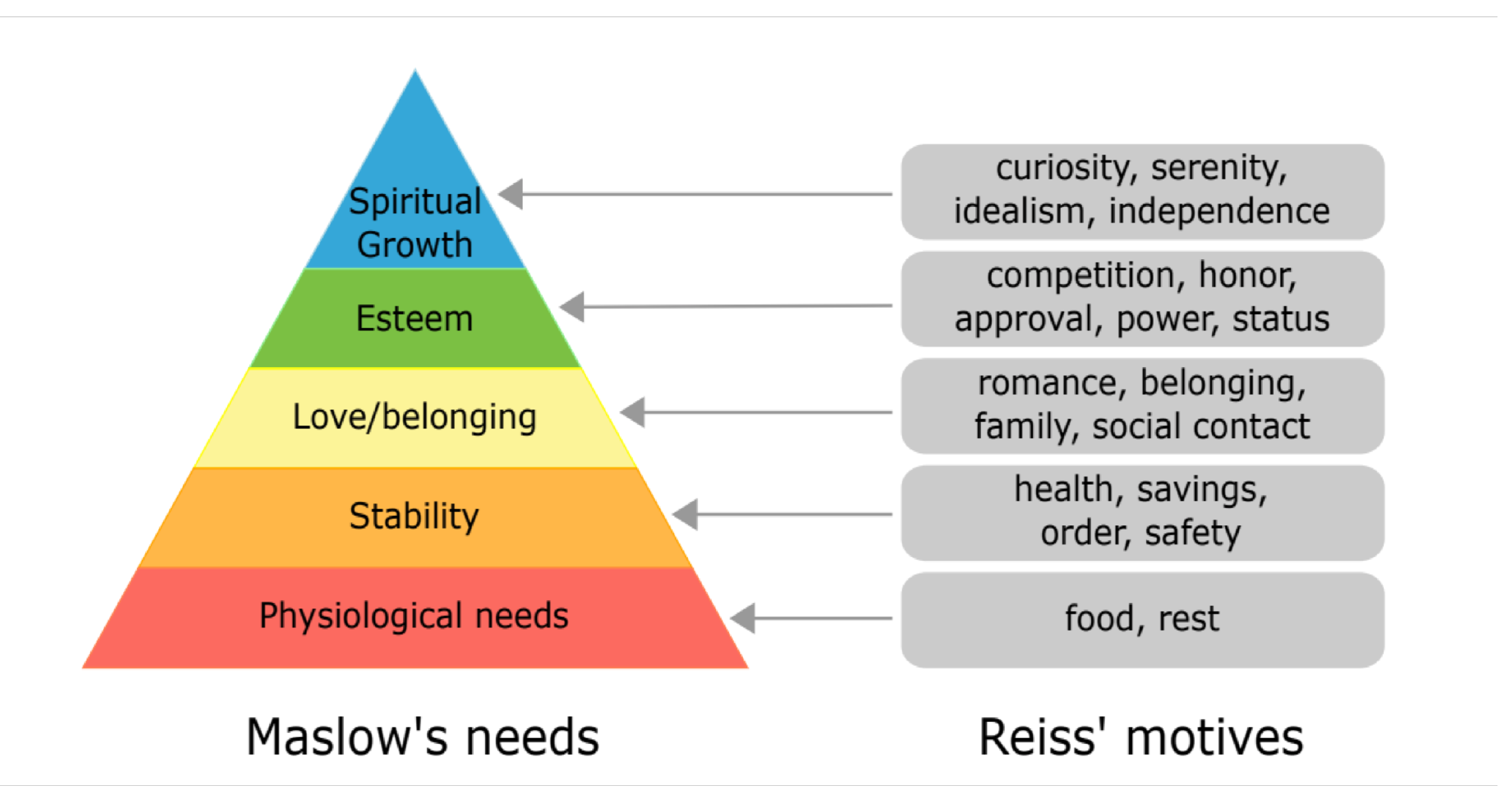}
    \caption{Maslow and Reiss Labels}
    \label{fig:labels}
\end{figure}

\begin{figure*}[ht!]
  \centering
    \includegraphics[scale=0.35]{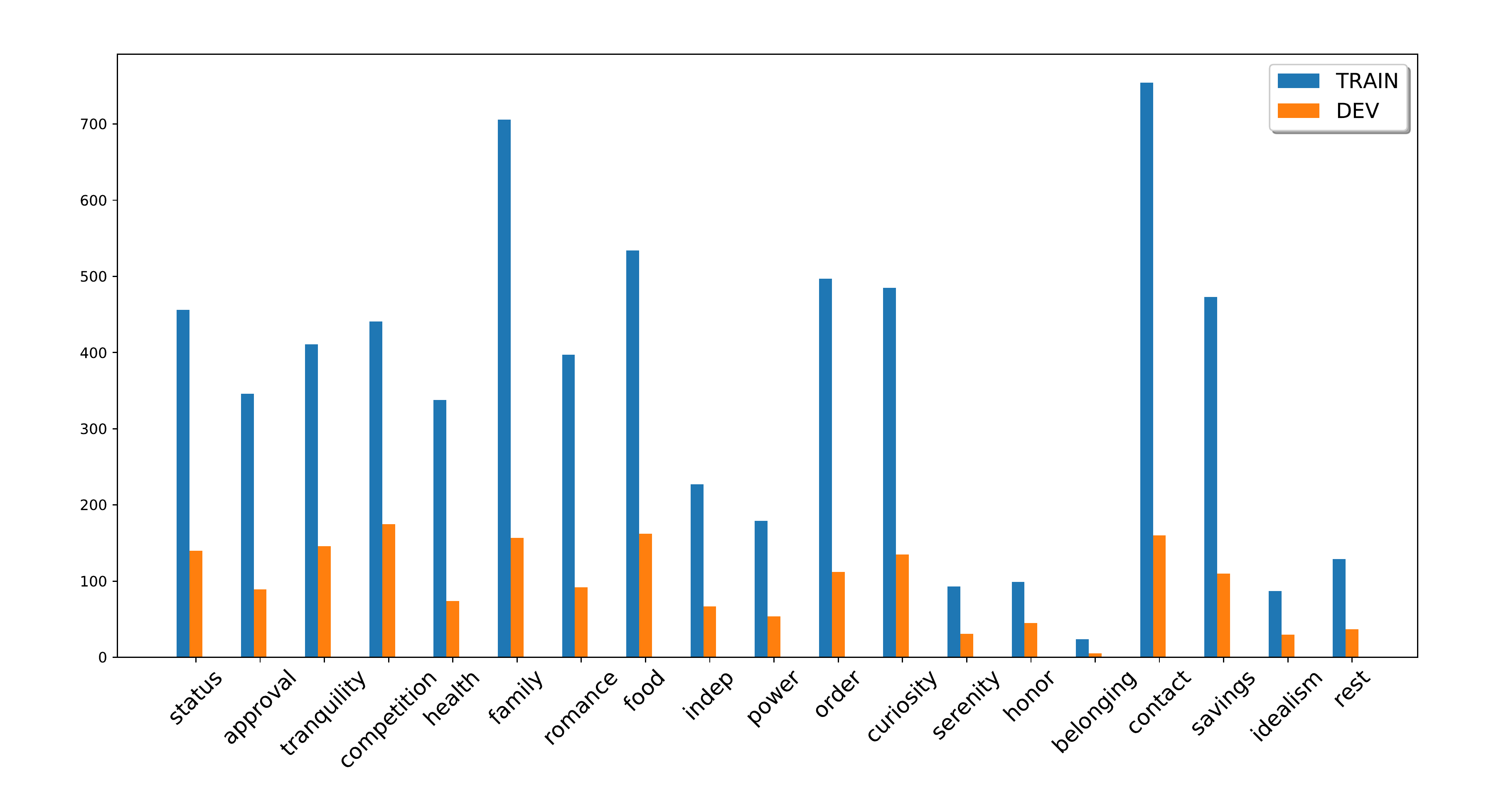}
    \caption{Train and Dev data statistics for Reiss Classification.}
    \label{fig:reiss_data}
\end{figure*}

\begin{figure*}[ht!]
  \centering
    \includegraphics[scale=0.29]{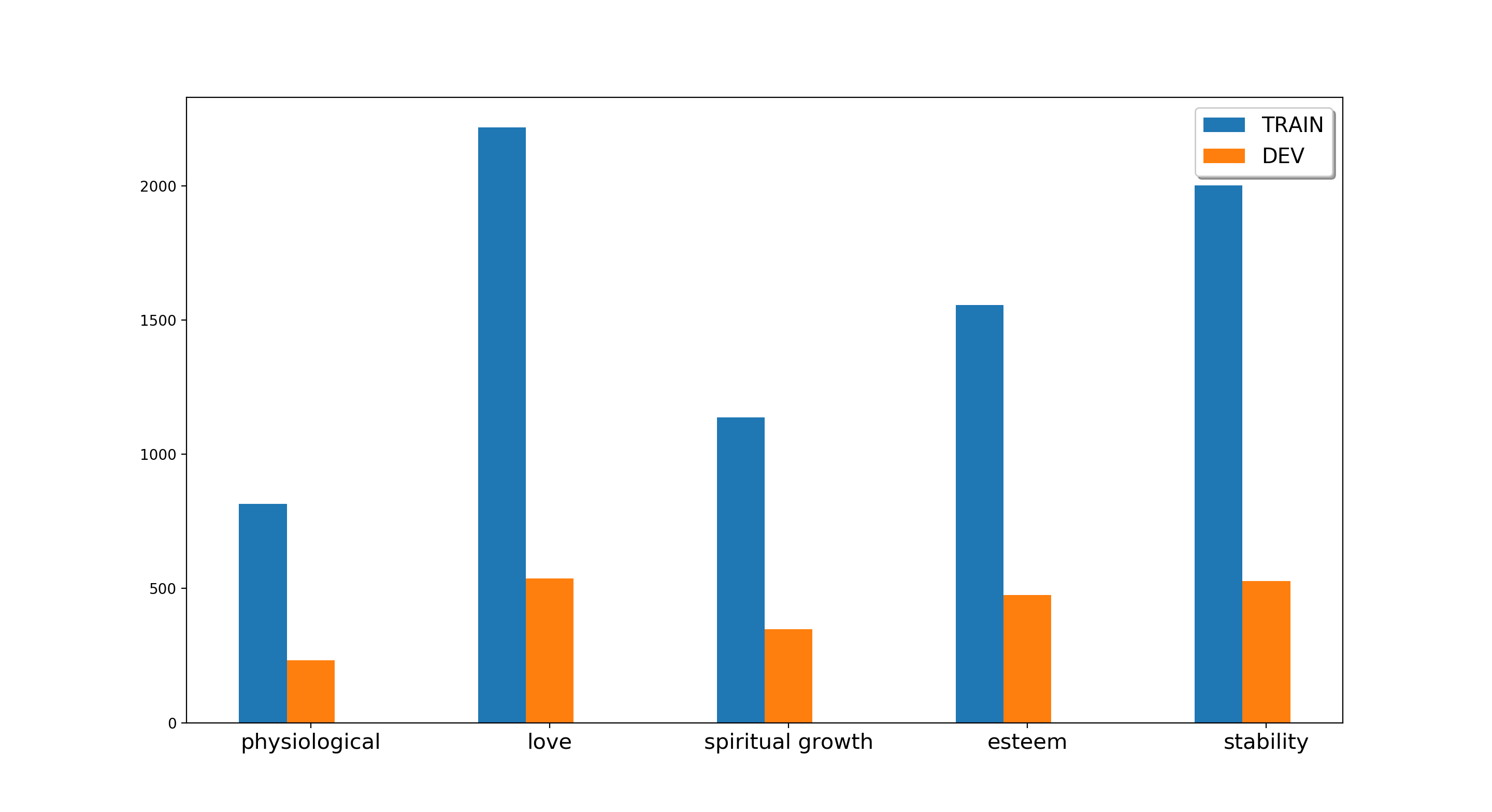}
    \caption{Train and Dev data statistics for Maslow Classification.}
    \label{fig:maslow_data}
\end{figure*}

We train and test our model on the \textit{Modeling Naive Psychology of Characters in Simple Commonsense Stories} dataset \cite{rashkin2018modeling}. It contains narrative stories where each sentence is annotated with a character and a set of human need categories from two inventories: Maslow's (with five coarse-grained) and Reiss's (with 19 fine-grained) categories. Figure \ref{fig:labels} portraits the labels in Reiss and Maslow and their relation.  Figures \ref{fig:reiss_data} and \ref{fig:maslow_data} depict the data distribution for the training and dev set for Reiss and Maslow respectively. As in prior work we select the annotations that display the ``majority label" i.e., categories voted on by \(\geq 2\) workers. Since no training data is available, similar to prior work we use a portion of the devset as training data, by performing a random split, using 80\% of the data to train the classifier, and 20\% to tune parameters. We use ConceptNet version 5.6.0 to extract commonsense knowledge. 

\begin{figure}[ht!]
  \centering
    \includegraphics[scale=0.5]{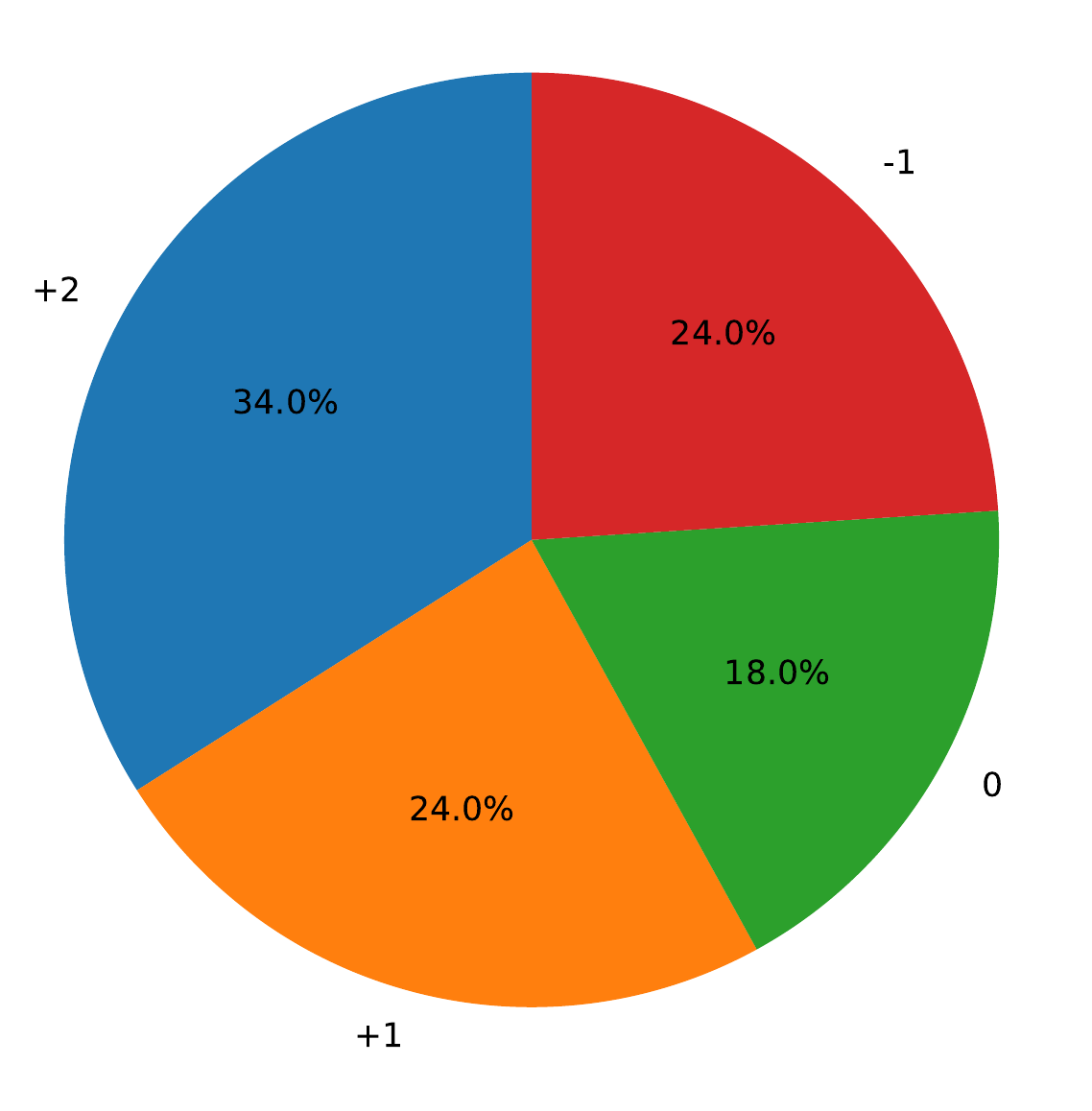}
    \caption{Human evaluation: Distribution of scores.}
    \label{fig:human_evaluation}
\end{figure}

\subsection{Training Details}
In training, we minimize the weighted binary cross entropy loss to train our multi-label classifier with the Adam optimizer \cite{kingma2014adam} with an initial learning rate of 0.001, a dropout-rate of 0.5 (dropout is applied to the input of each LSTM layer) and batch size of 32. We use 300 dimensional word embeddings and a hidden size of 100 for all Dense Layer and k = 3 for the selection of
top ranked paths. For Maslow labels, we use L2 regularization with \(\lambda = 0.01\), For Reiss labels, we use L2 regularization with \(\lambda = 0.1\). 

\subsection{Concept to Human Needs}
We manually aligned the human need categories to concepts in ConceptNet.
We used the name of the human needs to map them to identically named concepts from ConceptNet, except for 3 human needs classes,  which are as follows (Table \ref{tab:hn}): 
\begin{table}[ht!]
\small
\centering
\scalebox{0.99}{
\begin{tabular}{c c}
{\textbf{Concepts}}& {\textbf{Human needs}}\\
{tranquility} & {safety}\\
{serenity} & {calm}\\
{contact} & {social}\\
\end{tabular}
}
\caption{Concepts corresponding to Human needs}\label{tab:hn}
\end{table}

For Maslow's labels we use the 
mapping
for Reiss, as Maslow's categories are a subset of the Reiss categories, as shown in  Figure \ref{fig:labels}.

\subsection{Human evaluation}
We conduct human evaluation to test the effectiveness and relevance of the extracted commonsense knowledge paths. We randomly selected 50 sentence-context pairs with their gold labels from the dev set and extracted knowledge paths that contain the gold label (using CC+PPR for ranking). We asked three expert evaluators to decide whether the paths provide 
relevant 
information about the missing links between the concepts in the sentence and the human need (gold label). 
We asked them to 
assign scores according to the following definitions:\\ 
\begin{description}
\item[+2:] 
the path specifies
perfectly relevant information
to provide
the missing link between the concepts in the sentence and the human need.
\item[+1:]  the path contains a sub-path that specifies
relevant information to 
provide
the missing links between the concepts in the sentence and the human need.
\item[0:]
when the path is irrelevant but the starting and the ending nodes stand in a relation that is relevant to link the sentence and the expressed human need. (In this case,
either the path selected by our algorithm is not relevant or there is no relevant path connecting the nodes given the context.)
\item[-1:] 
the path is completely irrelevant.
\end{description}

Figure \ref{fig:human_evaluation} depicts the distribution of assigned scores (based on the majority class).
It shows that in 34\% of the cases 
our algorithm was able to select a relevant commonsense path. In another 24\% of cases a sub-part of the selected path was still considered relevant.

\subsection{Model Analysis and Visualization}
We study the visualization of attention distributions produced by our model. We provide examples for different scenarios. 
Here we show the results found by our best model i.e., BiLSTM+Self-Attention+Gated-Knowledge with CC+PPR as path selection method. 
\begin{figure*}[t]
\textbf{Case 1: Inclusion of knowledge path improves the performance when there is no context.\\}
\centering
\minipage{0.50\textwidth}
  \small{\textbf{Context:} No Context} \\
  \small{\textbf{Sentence:} Tina was out for a walk in the street.}\\
  \small{\textbf{True Label:} \textit{Health}} \\
  \small{\textbf{Predicted without Knowledge:} \textit{Serenity}}\\
  \small{\textbf{Predicted with Knowledge :} \textit{Health}} \\
\endminipage\hfill
\vspace*{-2mm}
\minipage{0.4\textwidth}
  \centering
  \includegraphics[width=8.0cm]{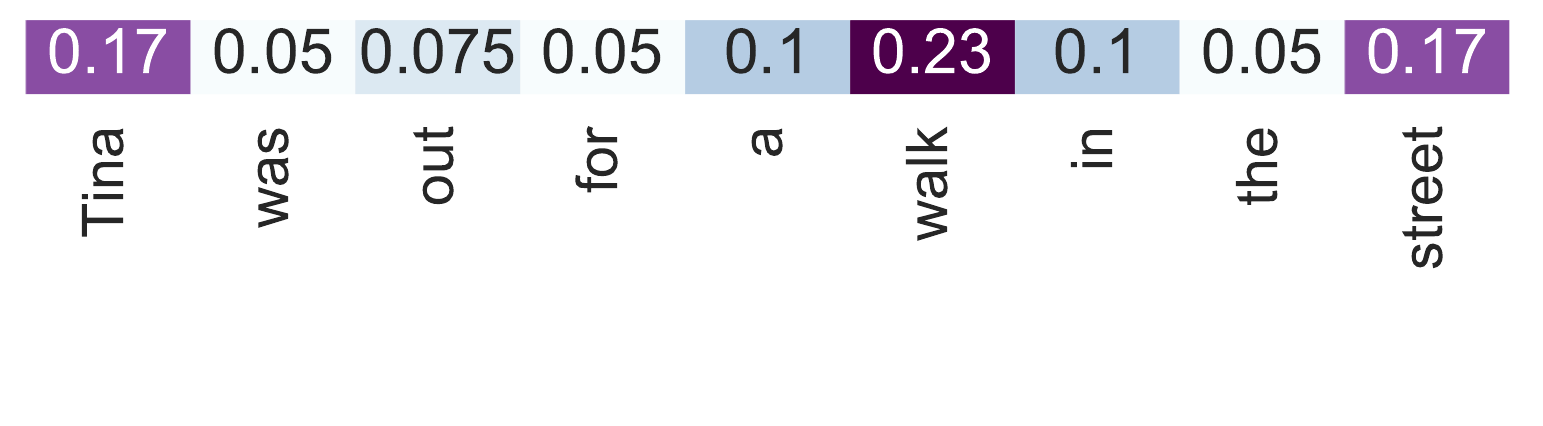}
\endminipage\hfill
\vspace*{-1mm}
\minipage{0.6\textwidth}
  \centering
  \includegraphics[width=\linewidth,height=5.0cm]{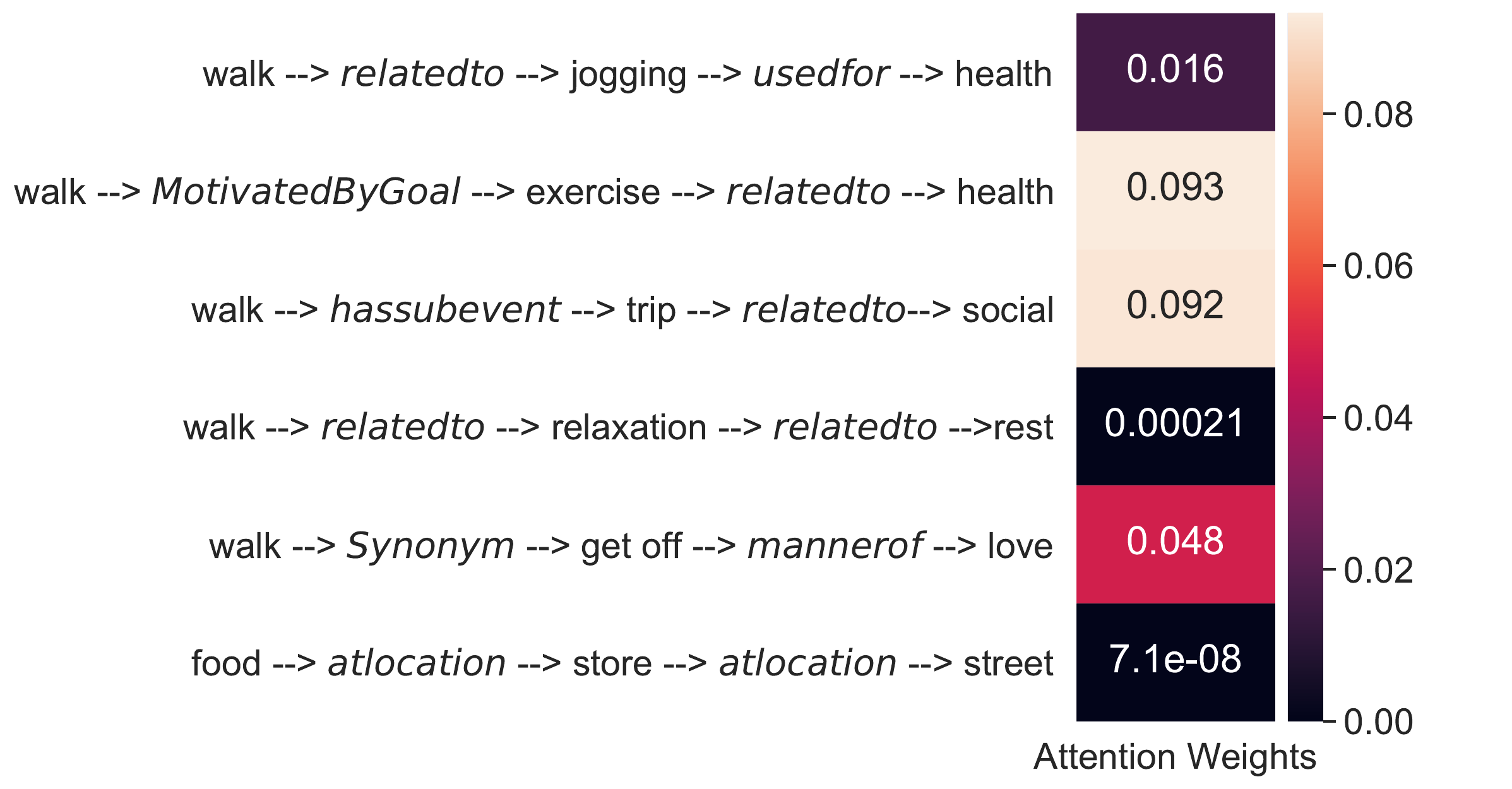}
\endminipage\hfill
\caption{Example 1: Visualizing the attention weights of the input sentence and of selected commonsense paths.}
\label{fig:attention_1}
\end{figure*}

\begin{figure*}[t]

\textbf{Case 2: Inclusion of knowledge paths improves the precision of the model.\\}

 \centering
\minipage{0.50\textwidth}

  \small{\textbf{Context:} No Context} \\
  \small{\textbf{Sentence:} Noah wanted to play golf against Nick.}\\
  \small{\textbf{True Label:} \textit{Competition}}\\
  \small{\textbf{Predicted without Knowledge:} \textit{Competition, Curiosity}}\\
  \small{\textbf{Predicted with Knowledge :} \textit{Competition}}
  
\endminipage\hfill
\vspace*{-2mm}
\minipage{0.35\textwidth}
  \centering
  \includegraphics[width=8.0cm]{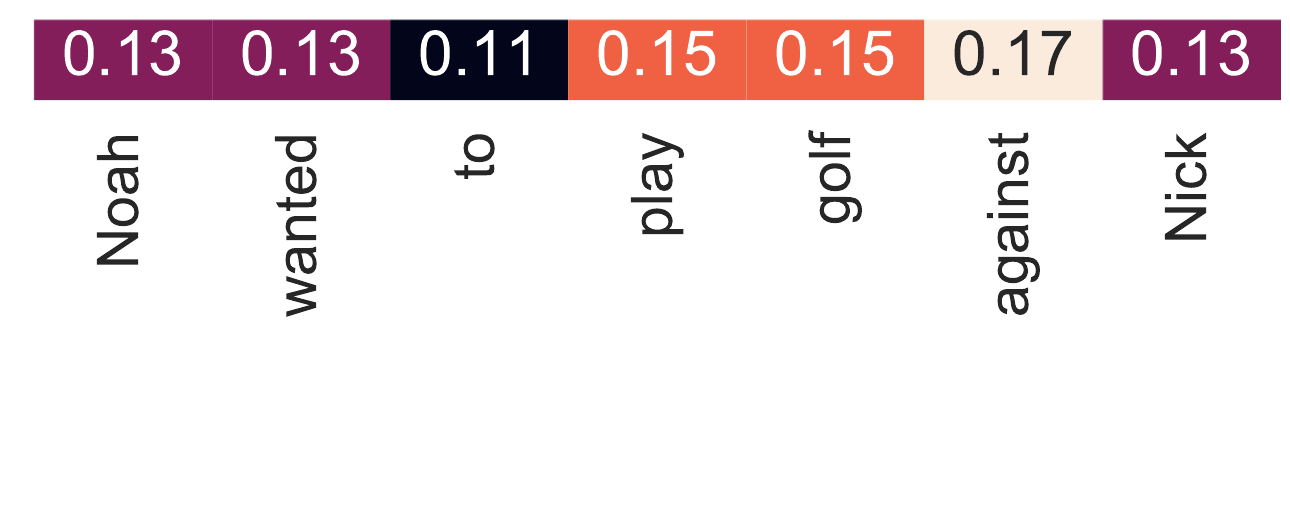}
\endminipage\hfill
\vspace*{-2mm}
\minipage{0.6\textwidth}
  \centering
  \includegraphics[width=\linewidth,height=5cm]{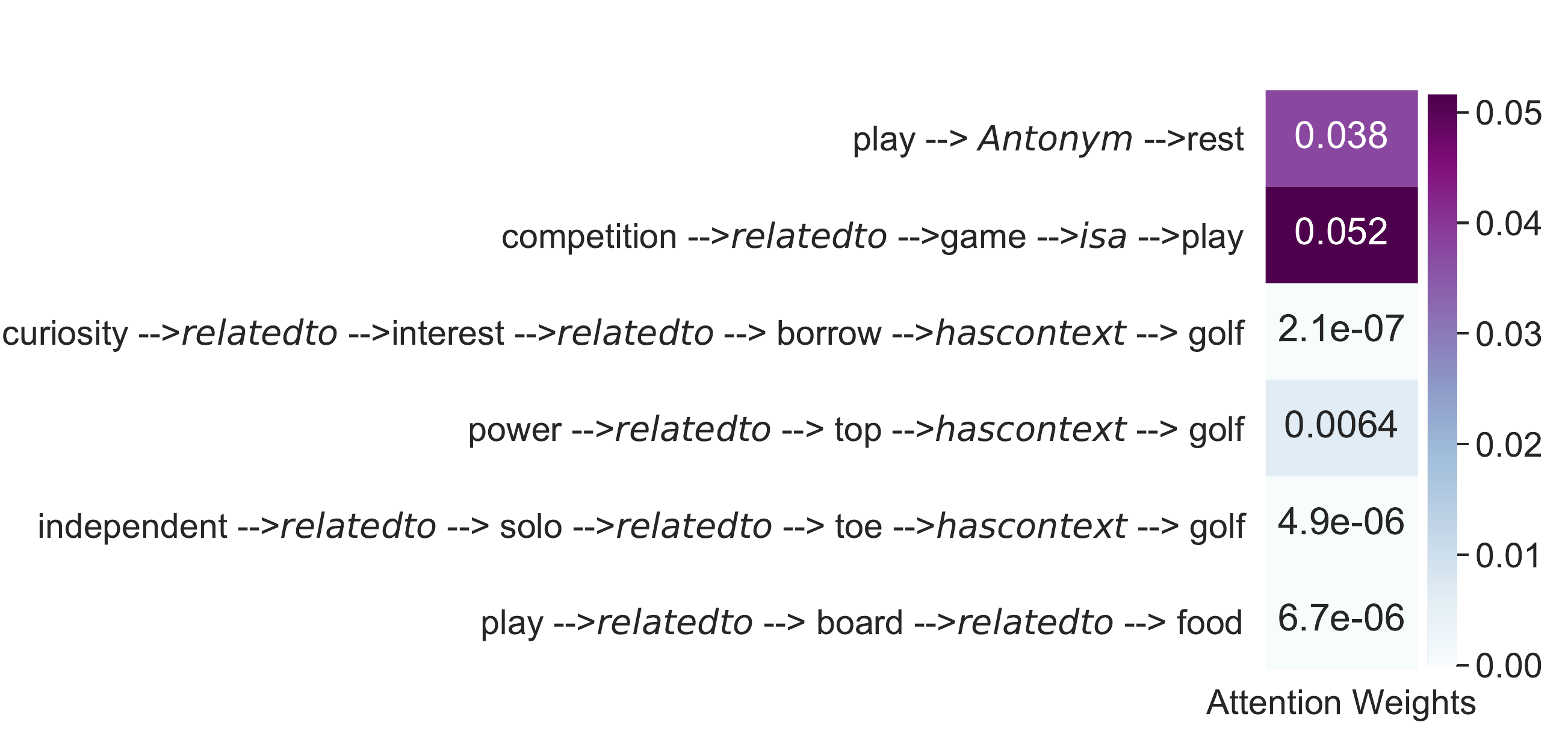}
\endminipage\hfill
\caption{Example 2: Visualizing the attention weights of the input sentence and of selected commonsense paths.}
\label{fig:attention_2}
\end{figure*}
\begin{figure*}[t]

\textbf{Case 3: Inclusion of knowledge paths improves the recall of the model\\}

\centering

\minipage{0.50\textwidth}
  \small{\textbf{Context:} Liv was a budding artist and she loved painting. She wanted to go to art classes, but her school didn't offer any!, So Liv got together with her friends and began brainstorming. They decided to form their own art group at the high school.} \\ 
  \small{\textbf{Sentence:} They made an after-school art club and named Liv president!}\\
  \small{\textbf{True Label:} \textit{Independent, Curiosity, Contact}}\\
  \small{\textbf{Predicted without Knowledge:} \textit{Contact}}\\
  \small{\textbf{Predicted with Knowledge :} \textit{Independent, Curiosity, Contact}}\\
\endminipage\hfill
\vspace*{-2mm}
\minipage{0.35\textwidth}
  \centering
  \includegraphics[width=8.0cm]{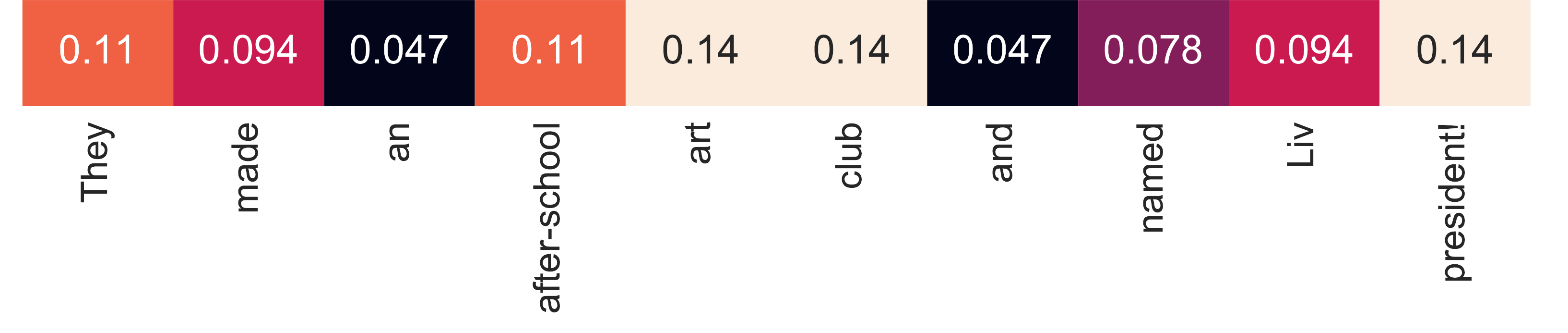}
\endminipage\hfill
\vspace*{-1mm}
\minipage{0.6\textwidth}
  \centering
  \includegraphics[width=\linewidth,height=6cm]{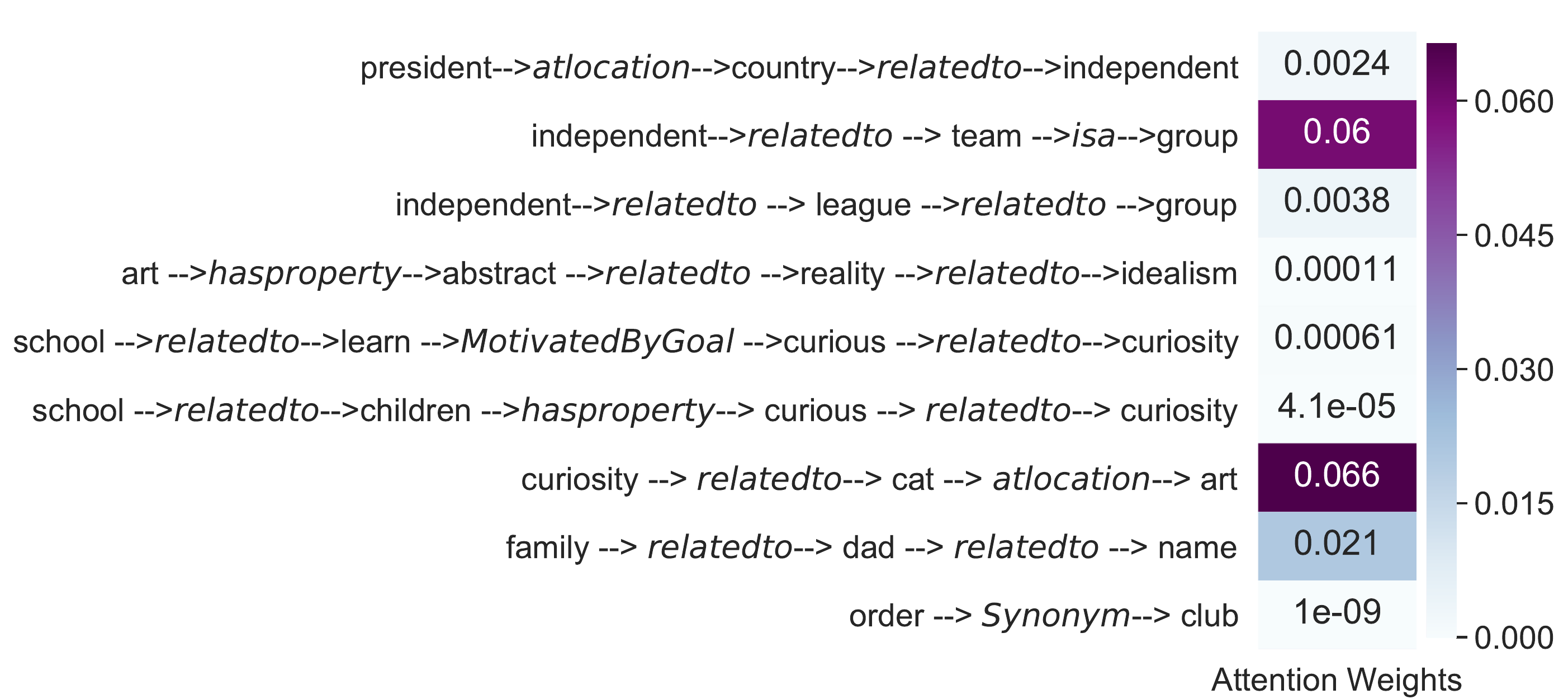}
\endminipage\hfill
\caption{Example 3: Visualizing the attention weights of the input sentence and of selected commonsense paths.}
\label{fig:attention_3}
  
\end{figure*}
\begin{figure*}[t]
\textbf{Case 4: In this case our model fails to attend to the relevant path. Although the graph-based ranking and selection algorithm were able to extract a relevant knowledge path, the neural model fails to correctly pick (attend to) the correct path.\\} 

 \centering
\minipage{0.50\textwidth}
  \small{\textbf{Context:} Tom was driving his car. He wanted to take a scenic way home. He deliberately passed his exit. Tom saw many beautiful trees.} \\ \small{\textbf{Sentence:} Tom took the scenic way home.}\\
  \small{\textbf{True Label:} \textit{Serenity}}\\
  \small{\textbf{Predicted without Knowledge:} \textit{Independent, Curiosity}}\\
  \small{\textbf{Predicted with Knowledge :} \textit{Family, Independent, Curiosity, Serenity}}\\
\endminipage\hfill
\vspace*{-2mm}
\minipage{0.35\textwidth}
  \centering
  \includegraphics[width=8.0cm]{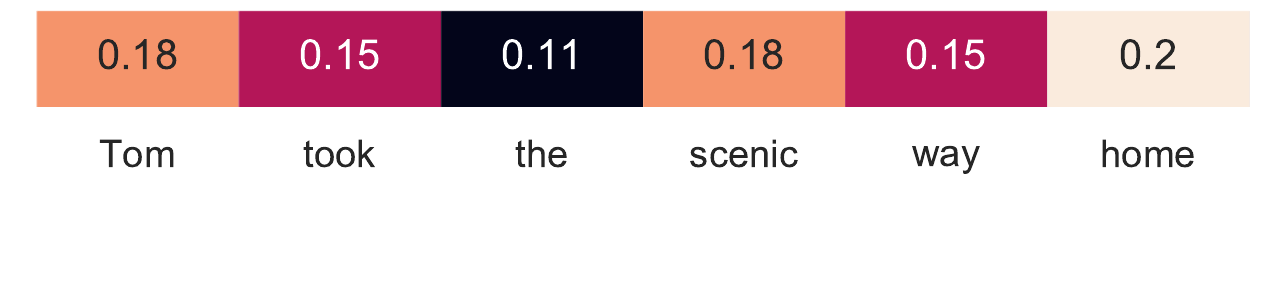}
\endminipage\hfill
\vspace*{-2mm}
\minipage{0.6\textwidth}
  \centering
  \includegraphics[width=\linewidth,height=6cm]{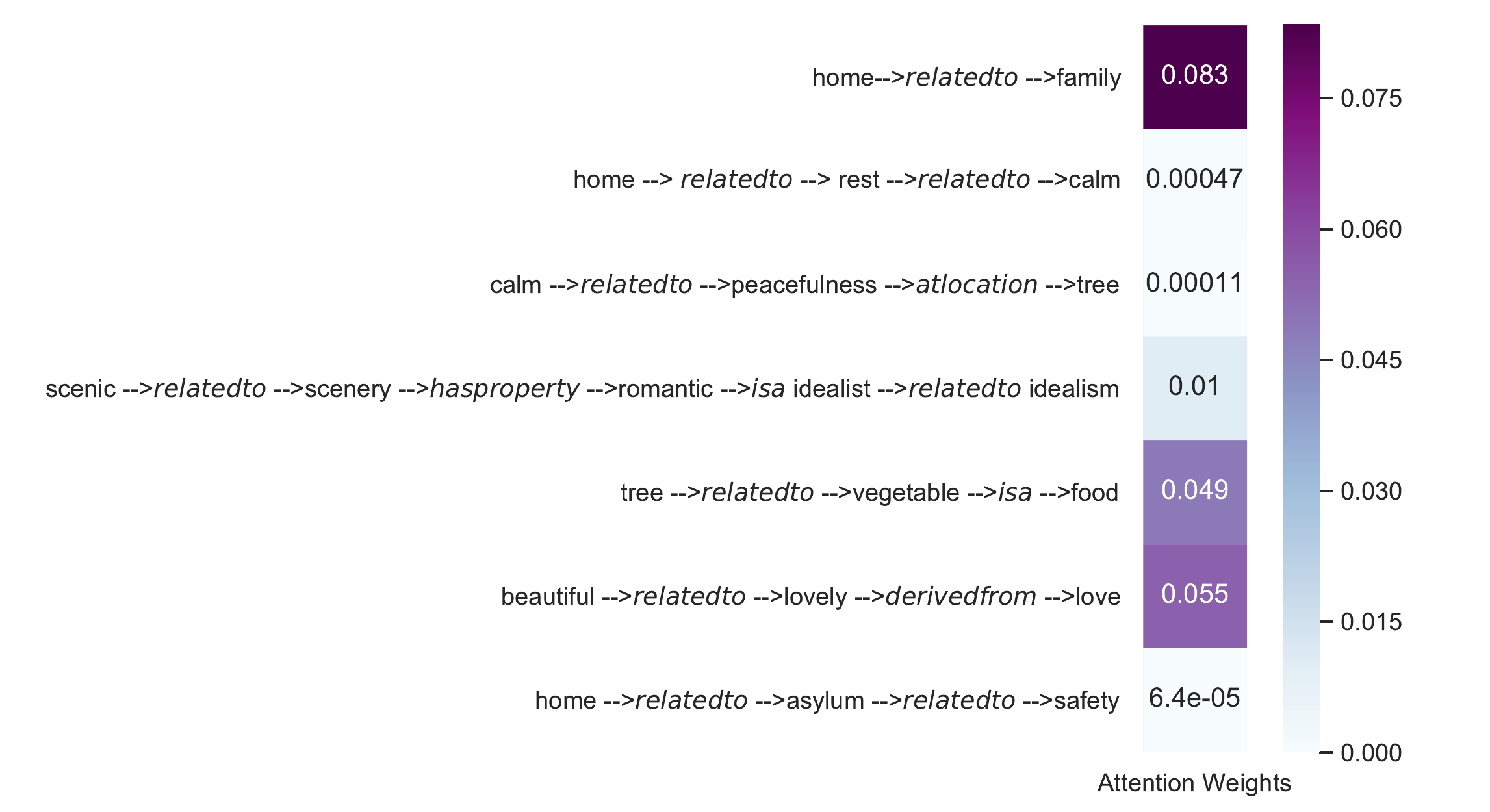}
\endminipage\hfill
\caption{Example 4: Visualizing the attention weights of the input sentence and of selected commonsense paths.}
\label{fig:attention_4}
\end{figure*}

\end{document}